\documentclass{article}

\usepackage[preprint]{neurips_2025}

\usepackage[utf8]{inputenc} 
\usepackage[T1]{fontenc}    
\usepackage{hyperref}       
\usepackage{url}            
\usepackage{booktabs}       
\usepackage{amsfonts}       
\usepackage{nicefrac}       
\usepackage{microtype}      
\usepackage[dvipsnames]{xcolor,colortbl}         
\usepackage{soul}
\usepackage{fontawesome} 

\newcommand{\hlc}[2][yellow]{{%
    \colorlet{foo}{#1}%
    \sethlcolor{foo}\hl{#2}}%
}

\usepackage{graphicx}
\usepackage{amsmath}
\usepackage{cleveref}
\usepackage{enumitem}
\usepackage{tabularx}
\usepackage{subcaption}
\usepackage{xspace}
\usepackage{wrapfig}
\usepackage{multirow}
\usepackage{arydshln} 

\usepackage[breakable,skins]{tcolorbox}
\usepackage{listings}
\Crefname{section}{Sec.}{Secs.}
\Crefname{equation}{Eq.}{Eqs.}
\Crefname{figure}{Fig.}{Figs.}
\Crefname{tabular}{Tab.}{Tabs.}
\Crefname{appendix}{App.}{Apps.}
\crefname{lstlisting}{prompt}{prompts}
\Crefname{lstlisting}{Prompt}{Prompts}

\title{J4R: Learning to Judge with Equivalent Initial State Group Relative Policy Optimization}

\author{%
    Austin Xu, Yilun Zhou, Xuan-Phi Nguyen, Caiming Xiong, Shafiq Joty\\[0.3em] 
  Salesforce AI Research\\
  Correspondence: \texttt{austin.xu@salesforce.com} \\
}

\newtoggle{comments}
\toggletrue{comments}
\togglefalse{comments}

\iftoggle{comments}{
    \newcommand{\yilun}[1]{\textcolor{blue}{(Yilun: #1)}}
    \newcommand{\austin}[1]{\textcolor{orange}{(Austin: #1)}}
    \newcommand{\xphi}[1]{\textcolor{red}{(Phi: #1)}}
    \newcommand{\shafiq}[1]{\textcolor{cyan}{(shafiq: #1)}}
}{
    \newcommand{\yilun}[1]{}
    \newcommand{\austin}[1]{}
    \newcommand{\xphi}[1]{}
    \newcommand{\shafiq}[1]{}
}

\newcommand{\benchmark}{ReasoningJudgeBench}
\newcommand{\judgeacc}{J4R}
\newcommand{\judgefull}{Judge for Reasoning (J4R)}
\newcommand{\judge}{J4R-CJ-7B}
\newcommand{\judgeins}{J4R-Qwen-Inst-7B}
\newcommand{\method}{EIS-GRPO\xspace}
\newcommand{\methodfull}{Equivalent Initial State GRPO\xspace}
\newcommand{\methodfullfull}{Equivalent Initial State Group Relative Policy Optimization\xspace}

\begin{document}

\maketitle

\begin{abstract}
To keep pace with the increasing pace of large language models (LLM) development, model output evaluation has transitioned away from time-consuming human evaluation to automatic evaluation, where LLMs themselves are tasked with assessing and critiquing other model outputs. LLM-as-judge models are a class of generative evaluators that excel in evaluating relatively simple domains, like chat quality, but struggle in reasoning intensive domains where model responses contain more substantive and challenging content. To remedy existing judge shortcomings, we explore training judges with reinforcement learning (RL). We make three key contributions: (1) We propose the Equivalent Initial State Group Relative Policy Optimization (\method) algorithm, which allows us to train our judge to be robust to positional biases that arise in more complex evaluation settings.  
(2) We introduce \benchmark{}, a benchmark that evaluates judges in diverse reasoning settings not covered by prior work. 
(3) We train \judgefull{}, a 7B judge trained with \method that outperforms GPT-4o and the next best small judge by 6.7\% and 9\%, matching or exceeding the performance of larger GRPO-trained judges on both JudgeBench and \benchmark{}.
\end{abstract}

\begin{center}
\faGithub: \href{https://github.com/SalesforceAIResearch/ReasoningJudgeBench}{GitHub} \hspace{5mm} \faBook: \href{https://huggingface.co/datasets/Salesforce/ReasoningJudgeBench}{ReasoningJudgeBench}
\end{center}

\section{Introduction}
Large language models (LLMs) and LLM-powered agentic systems have been tasked with solving increasingly difficult problems that require strong reasoning abilities~\citep{ke2025survey}. As LLMs are deployed in reasoning intensive-tasks, evaluation methods have come under mounting pressure to efficiently and accurately evaluate and critique LLM outputs, especially when the outputs are not easily verifiable with rules. While human evaluation remains the gold-standard in such cases, automatic evaluation approaches, whether it be scalar reward models (RMs) or LLM-as-judges, 
have been deployed as a more scalable alternative. However, these approaches have shortcomings~\citep{cemri2025multi}: Current verifiers tend to lock onto stylistic details over substance~\citep{feuer2024style,liu2024rm,zeng2023evaluating}. RMs are incapable of explaining their decisions, while judge models are prone to various biases such as positional~\citep{li2023generative,wang2023large} and length bias~\citep{zeng2023evaluating,park2024offsetbias}. In all, these drawbacks lead to evaluators that struggle to evaluate other models in reasoning settings~\citep{tan2024judgebench,zhou2025evaluating}. 

\begin{figure}
    \centering
    \includegraphics[width=\linewidth]{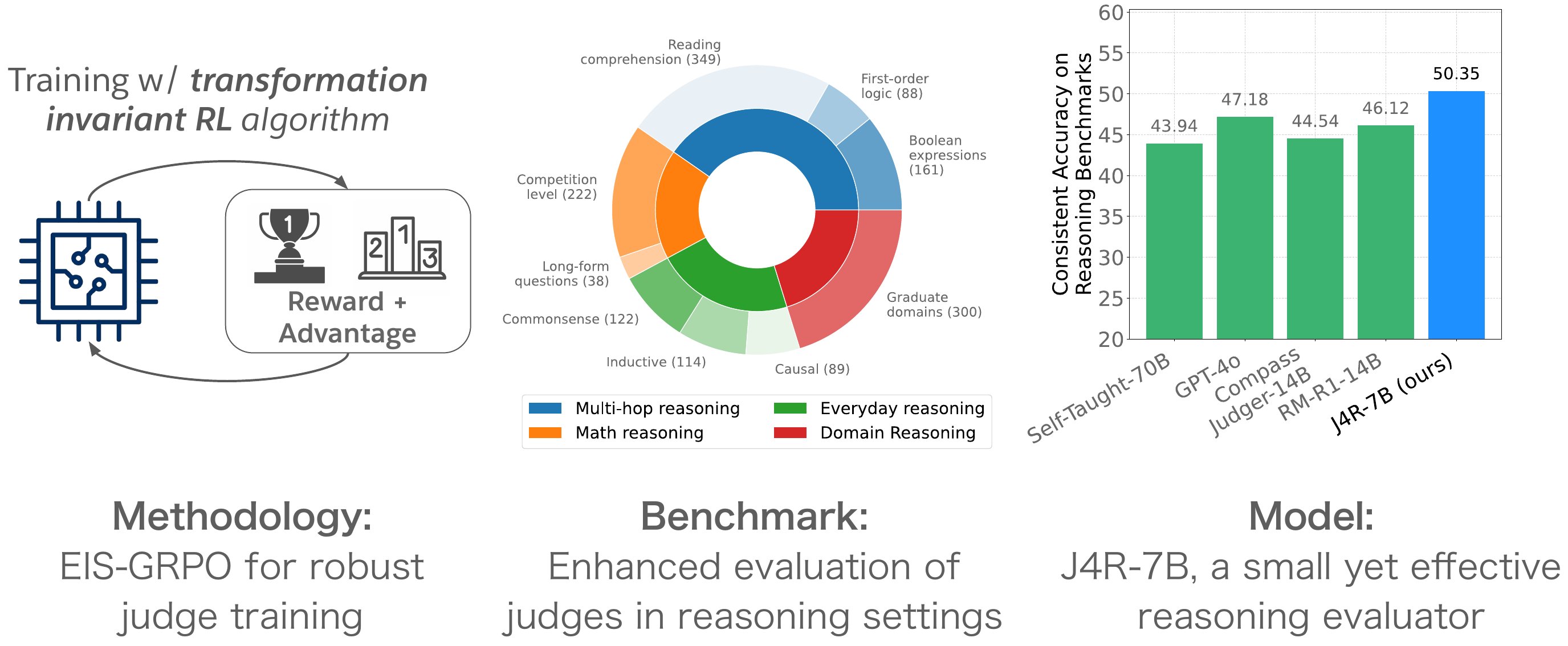}
    \caption{Overview of our three contributions: (1) We propose EIS-GRPO, a RL training algorithm for training positionally robust judge models. (2) We create ReasoningJudgeBench, a benchmark of 1,483 pairwise samples across diverse and underexplored reasoning settings. (3) We train J4R-7B, a lightweight yet high-performing judge specifically for reasoning evaluation.}
    \label{fig:overview}
\end{figure}

This work explores using reinforcement learning from verifiable rewards (RLVR)~\citep{lambert2024t,shao2024deepseekmath,guo2025deepseek} to improve the ability of evaluators, specifically LLM-as-judge models, in evaluating model responses for reasoning problems. Automatic evaluation is a natural setting for RLVR: Reinforcement learning (RL) allows the model to learn how to generate chain-of-thought (CoT) critiques without supervision from distilled CoT critiques, which are commonly used in judge model training (e.g.,~\cite{kim2023prometheus,kim2024prometheus,wang2024direct,ye2024beyond}). Furthermore, automatic evaluation outputs have a limited set of final outcomes (e.g., ``A'' or ``B'' for pairwise comparisons, 1-5 for single ratings), making them naturally ``verifiable''.

Before simply applying standard RL algorithms~\citep{shao2024deepseekmath,guo2025deepseek} with judge data, we first take a step back and carefully look into where judges fail in reasoning settings. Our analysis in~\Cref{sec:method:judge} shows that as task difficulty grows, existing judge models, including those trained with RL, suffer from increasing \textit{positional inconsistency}, where swapping the order of the candidates does not lead to an equivalent swap in judgment. This inconsistency indicates that the judge is randomly guessing, rather than substantively assessing. To remedy this, we propose \methodfullfull (\method), which explicitly trains the model to treat transformations of the same initial state (i.e., input context) as equivalent, instilling consistency and robustness in judges with zero additional training overhead. Using \method, we train \judgefull, a 7B LLM judge that specializes in evaluating reasoning. 
J4R outperforms GPT-4o by 6.7\% and approaches the performance of 32B RL-trained judges on a suite of reasoning-focused judge benchmarks.

In our evaluation, we find existing benchmarks of reasoning judges lack breadth. Notably, JudgeBench~\citep{tan2024judgebench} consists only of 350 pairwise samples sourced from MMLU-Pro~\citep{wang2024mmlu}, LiveBench~\citep{white2024livebench}, and LiveCodeBench~\citep{jain2024livecodebench}, missing coverage of different, uniquely challenging reasoning tasks, like inductive or commonsense reasoning.
This leads us to create \benchmark{} to augment existing benchmarks with new, challenging types of reasoning. As shown in~\Cref{fig:overview}, our contributions are threefold:
\begin{itemize}[leftmargin=*,noitemsep,topsep=5pt]
    \item \textbf{Methodology} (\Cref{sec:method}): We propose \methodfull, a simple but effective data augmentation method for GRPO that allows us to combat judge positional bias while \textit{improving} performance over data-driven approaches.
    \item \textbf{Benchmark} (\Cref{sec:benchmark}): We create \benchmark{}, a benchmark of 1,483 challenging pairwise samples. 
    Compared to existing reasoning-focused judge benchmarks, \benchmark{} contains 4x more samples 
    covering more diverse reasoning settings.
    \item \textbf{Model and Analysis} (\Cref{sec:exp}): Using \method, we train \judge{}, a 7B model that punches above its weight class: It matches the performance of larger judge models in reasoning evaluation and outperforms the next best $\leq14$B judge by 13\% and 9\% on JudgeBench and ReasoningJudgeBench.
    We also perform a suite of analyses focused on (1) the utility of RL versus other common judge training recipes, (2) choice of initial model, and (3) the efficacy of \method versus other bias-mitigation training strategies.
\end{itemize}

\begin{figure}
    \centering
    \includegraphics[width=0.73\linewidth]{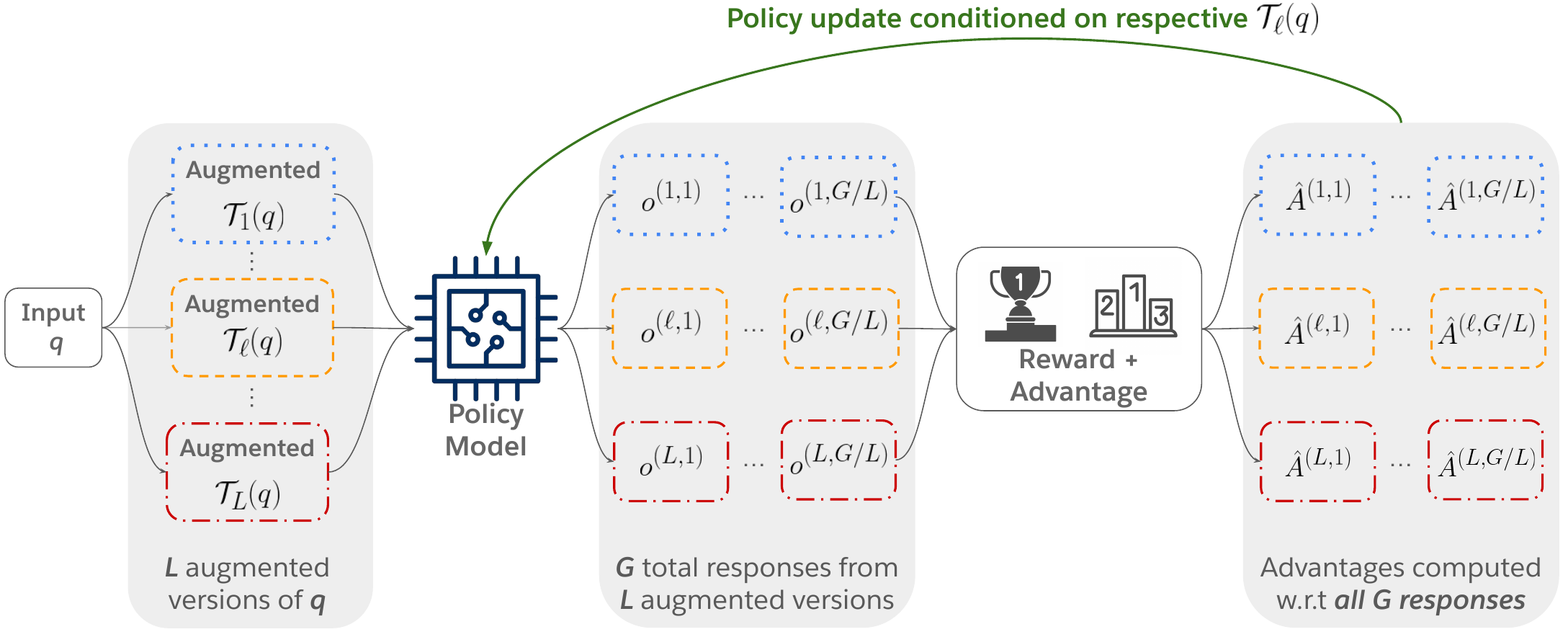}
    \caption{Overview of \methodfull{}: Each input $q$ is first transformed via $\mathcal{T}_\ell$ to a \textit{substantively equivalent} version of $q$ and used to sample a subgroup of responses. Responses are rewarded with outcome verification and advantages are computed jointly across subgroups and within subgroups before being used to update the policy model with respect to the transformed input. Joint subgroup advantage computation teaches the model to treat all transformed inputs as \textit{equivalent}, training the model to be robust to non-substantive transformations of the input.}
    \label{fig:eis_grpo}
    \vspace{-1\baselineskip}
\end{figure}

\section{Background and related work} 

\textbf{Learning to reason with reinforcement learning.} LLMs have been tasked with solving increasingly complex problems that require sound reasoning abilities. This lead to the development of long CoT models like OpenAI's o1~\citep{jaech2024openai} or DeepSeek-R1~\citep{guo2025deepseek}, which expend large amounts of inference-time compute generating ``thoughts'' before answering. DeepSeek-R1 revealed that \textit{reinforcement learning} using rule-based verification as a reward signal, broadly known as RL from Verifiable Rewards~\citep{lambert2024t}, is capable of unlocking powerful reasoning abilities from pretrained LLMs. As a result, RL and long CoT distillation have emerged as popular training paradigms for reasoning enhancement~\citep{yu2025dapo,liu2025understanding,muennighoff2025s1}; See~\cite{chen2025towards,zeng2024scaling,xu2025towards} for a complete overview.

\textbf{Automatic evaluation.} Advances in reasoning have largely focused on settings where answers are easily verifiable (e.g., math). However, many complex problems, such as long-form report generation, cannot be graded solely on answer correctness. As a result, strong automatic evaluators are needed for such reasoning tasks. Automatic evaluation methods broadly fall under two paradigms: reward models (RMs)~\citep{zhong2025comprehensive}, which are non-generative evaluators that output scaler scores, and LLM-as-judge~\citep{gu2024survey}, which are generative. The latter has gained popularity recently with a line of work training specialized judge models using supervised finetuning (SFT)~\citep{li2023generative,kim2023prometheus,kim2024biggen}, preference optimization~\citep{wang2024direct,ye2024beyond,hu2024themis,wang2024self,saha2025learning}, and hybrid objectives using generation and classification losses~\citep{zhang2024generative,mahan2024generative}. 

Only recently have methods for improving reasoning been applied to judges, with~\cite{liu2025inference} exploring inference-time scaling. The two most related works to ours are JudgeLRM~\citep{chen2025judgelrm} and the concurrent RM-R1~\citep{chen2025rmr1rewardmodelingreasoning}. JudgeLRM conducts an initial study of applying Group Relative Policy Optimization (GRPO) for training judge models, finding that judges trained with RL outperform those trained with SFT. However, their evaluation is limited to two non-reasoning benchmarks. RM-R1 fixes the training algorithm, GRPO, and focuses on analyzing various parts of the RL training pipeline (e.g., such as the impact of SFT warm-up). Our work can be seen as \textit{complementary} to RM-R1: We propose a novel training \textit{methodology}, \method{}, which was developed after analyzing key failure modes of judges in reasoning settings, to which many of the same findings in RM-R1 may apply. More broadly, in contrast to past advances in judge training, which simply apply the latest training algorithms (DPO, GRPO, etc.) for judges, we propose a methodological advancement: To our knowledge, our work is the \textit{first RL-based training algorithm designed with training judges in mind}.

Our analysis in~\Cref{sec:method:judge} reveals that using RL training alone is insufficient for evaluating in more difficult reasoning domains. Concretely, when evaluating challenging samples, RL-trained judges still exhibit random guessing behavior, which manifests as pairwise \textit{inconsistency}: When the order of responses is switched, the judge's chosen response switches. In the next section, we present a general-purpose algorithmic solution, \methodfull, which can be used to train judges to not only reason about responses in detail, but also to be immune to positional biases.

\section{\methodfull}\label{sec:method}
Here we present \methodfull{}, a policy gradient algorithm for training models to be robust to input transformations. Our presentation here is in a general manner, and we describe specific application to judge model training in~\Cref{sec:method:judge}. While what we present is specifically motivated via GRPO, the core ideas are generally applicable other algorithms, like RLOO~\citep{kool2019buy}.

\subsection{Background: Group Relative Policy Optimization}\label{sec:method:background}
In what follows, we use subscripts indexed with $t$ to denote token index, superscripts indexed with $(i)$ to denote sample index (i.e., if multiple outputs are generated for the same input).

\textbf{Proximal Policy Optimization (PPO)}~\citep{schulman2017proximal} is a policy gradient algorithm that constrains policy updates, preventing large updates and resulting in more stabilized training. When applied by earlier works~\citep{ouyang2022training,stiennon2020learning} for finetuning LLMs, given a question (prompt) $q$ , each output token $o_t$ is modeled as a separate \textit{action}, with $(q, o_{<t})$ as the \textit{state}. Specifically, given a dataset $\mathcal{D} = \{q\}$ of questions $q$, the LLM is optimized with the PPO objective
\small
\begin{align}
    \mathcal{L}_{\text{PPO}}(\theta) &= \mathbb{E}_{q\sim\mathcal{D}, o \sim \pi_{\theta_{\text{old}}}(\cdot|q)}
    \left[
    \frac{1}{|o|} \sum\limits_{t=1}^{|o|} \left(\min\left\{r_{t}(\theta)\hat{A}_{t}, \text{clip}\left(r_{t}(\theta); \epsilon\right)\hat{A}_{t}\right\} - \beta D_{\text{KL}}(\pi_{\theta} || \pi_{\text{ref}})\right)\right],
\end{align}
\normalsize
where $\epsilon$ is the clipping value that controls the allowable range of policy update, $\hat{A}_t$ is the \textit{advantage} estimate for time $t$, $D_{\text{KL}}$ is the KL divergence, which regularizes training by discouraging large deviation from a reference policy $\pi_{\text{ref}}$, and $r_t(\theta)$ and clip are defined as
\small
\begin{align}\label{eq:policy_ratio}
    r_{t}(\theta) = \frac{\pi_{\theta}(o_{t} | q, o_{<t})}{\pi_{\theta_{\text{old}}}(o_{t} | q, o_{<t})} \quad \text{and} \quad \text{clip}(x; \epsilon) =\min(1 + \epsilon, \max(1-\epsilon,x)),
\end{align}
\normalsize
respectively. The original implementation of PPO estimated the advantage with Generalized Advantage Estimation (GAE)~\citep{schulman2015high}, which uses a value function that is trained jointly with the policy model and a reward model, typically trained in advance with labeled preference data. At a very high level, PPO updates the model by computing how much better or worse a particular action $o_t$ is than average  via $\hat{A}_t$, then updates the model in a constrained region accordingly.

\textbf{Group Relative Policy Optimization (GRPO)}~\citep{shao2024deepseekmath} was proposed to simplify the PPO training pipeline, requiring instead a dataset $\mathcal{D} = \{(q,a)\}$ of question $q$ and answer $a$ pairs. GRPO considers an \textit{entire output} $o$ as a single \textit{action}, with original question $q$ being the \textit{(initial) state}. GRPO omits the reward model and value function by sampling $G$ \textit{complete responses} $o^{(1)}, \ldots, o^{(G)}$ per question $q$ (known as a \textit{group} of size $G$) from $\pi_{\theta_{\text{old}}}$, then computing the reward $R_i$ for each $o^{(i)}$ using rule-based verification with answer $a$. Then, the advantage $\hat{A}_{t}^{(i)}$ is computed as $\hat{A}_{t}^{(i)} = \left(R_i - \bar{R}\right)/\sigma_{R}$, where $\bar{R}$ and $\sigma_{R}$ denote the mean and standard deviation of the rewards $\{R_1, \ldots, R_G\}$. Above, the advantage does not depend on the time-step $t$ (i.e., each token is assigned the same advantage), and as such, we drop the subscript $t$ moving forward. With this advantage estimator, the GRPO objective is
\small
\begin{align}
    \mathcal{L}_{\text{GRPO}}(\theta) &= \mathbb{E}_{(q,a)\sim\mathcal{D}, \{o^{(i)}\}_{i=1}^G \sim \pi_{\theta_{\text{old}}}(\cdot|q)}\nonumber\\
    &\left[
    \frac{1}{G} \sum\limits_{i=1}^G \frac{1}{|o^{(i)}|} \sum\limits_{t=1}^{|o^{(i)}|} \left(\min\left\{r_t^{(i)}(\theta)\hat{A}^{(i)}, \text{clip}\left(r_t^{(i)}(\theta); \epsilon\right)\hat{A}^{(i)}\right\} - \beta D_{\text{KL}}(\pi_{\theta} || \pi_{\text{ref}})\right)\right],
\end{align}
\normalsize
where $r_t^{(i)}(\theta)$ is defined the same as $r_t(\theta)$ in~\Cref{eq:policy_ratio} for output $o^{(i)}$. Many variants of GRPO have recently been proposed, like DAPO~\citep{yu2025dapo} (using different values for high/low clipping, among other changes) and DR-GRPO~\citep{liu2025understanding} (removing $\sigma_R$ in the advantage and changing the loss normalization).

\subsection{\methodfull (\method): Augmenting with State Equivalence}\label{sec:method:esgrpo}

The vanilla formulation of GRPO considers any two questions $q$ and $q^\prime$ as different states regardless of their specific contents. However, within a real-world collection of inputs, some states may exhibit \textit{equivalence structure}~\citep{li2006towards}: Different questions may contain the same \textit{relevant} information while differing in \textit{irrelevant} ways (e.g., stylistic). Consider the following toy example in the math domain:
\small
\begin{align*}
    q = \text{``Solve \texttt{[problem]}. Think step-by-step.''} \quad \text{and} \quad q^\prime = \text{``Provide reasoning before answering: \texttt{[problem]}''}
\end{align*}
\normalsize
Above, $q$ and $q^\prime$ contain the same \textit{relevant} information, i.e., the objective is to solve \texttt{[problem]}, which requires the same reasoning process (actions) to solve. However, they differ in \textit{irrelevant} ways (e.g., CoT prompting language used). While identifying such equivalent problems in large-scale datasets can be difficult, \textit{transforming a given state (i.e., input) to a set of equivalent versions is easy} via hand-crafted data augmentation strategies (e.g., via paraphrasing).

Rather than sampling $G$ responses from $q$, we transform $q$ into $L$ equivalent versions, $\mathcal T_1(q), ..., \mathcal T_L(q)$, and then sample $G/L$ responses $\{o^{(1,\ell)}, \ldots, o^{(G/L, \ell)}\}$ from each $T_{\ell}(q)$, for a total of $G$ responses across all $\mathcal{T}_\ell(q)$\footnote{
    In practice, one can vary the number of outputs sampled for each equivalent initial state $T_\ell(q)$. However, for ease of notation and presentation, we assume fixed subgroup sizes of $G/L$.}. 
We call the set of responses sampled from a given $T_{\ell}(q)$ a \textit{subgroup}. Then, an \textcolor{orange}{\textit{advantage that is computed jointly across all subgroups}} is added to the subgroup-only advantage, and used to update the \textcolor{blue}{\textit{policy conditioned on each transformed state} $\mathcal{T}_\ell(q)$}. 
Concretely, suppose each output $o^{(i,\ell)}$ receives reward $R^{(i,\ell)}$, and denote $R^{[L]} = \{R^{(1,1)},\ldots,R^{(G/L,1)}, \ldots, R^{(1,L)},\ldots,R^{(G/L,L)}\}$ as the set of rewards across all subgroups and $R^\ell = \{R^{(1,\ell)},\ldots,R^{(G/L,\ell)}\}$ as the set of rewards within each subgroup. Then, define $\hat{A}^{(i,\ell)}$ and $r_t^{(i,\ell)}(\theta)$ as 
\small
\begin{align}\label{eq:esgrpo_ratio}
    \hat{A}^{(i,\ell)} =  (R^{(i,\ell)} - \textcolor{orange}{\bar{R}^{[L]}}) / \textcolor{orange}{\sigma_{R^{[L]}}} + (R^{(i,\ell)} - \bar{R}^{\ell}) / \sigma_{R^\ell}
    \quad \text{and} \quad 
    r_{t}^{(i,\ell)}(\theta) = \frac{\pi_{\theta}(o^{(i,\ell)}_{t} | \textcolor{blue}{\mathcal{T}_{\ell}(q)}, o^{(i,\ell)}_{<t})}{\pi_{\theta_{\text{old}}}(o_{t}^{(i,\ell)} | \textcolor{blue}{\mathcal{T}_{\ell}(q)}, o^{(i,\ell)}_{<t})}.
\end{align}
\normalsize
The \method objective takes the form
\small
\begin{align}\label{eq:esgrpo_objective}
    \mathcal{L}&_{\text{\method}}(\theta) = \mathbb{E}_{(q,a)\sim\mathcal{D}, \left\{\{o^{(i,\ell)}\}_{i=1}^{G/L} \sim \pi_{\theta_{\text{old}}}(\cdot|\mathcal{T}_{\ell}(q))\right\}_{\ell=1}^L}\nonumber\\
    &\left[
    \frac{1}{G} \sum\limits_{\ell=1}^L\sum\limits_{i=1}^{G_\ell} \frac{1}{|o^{(i,\ell)}|} \sum\limits_{t=1}^{|o^{(i,\ell)}|} \left(\min\left\{r_{t}^{(i,\ell)}(\theta)\hat{A}^{(i,\ell)}, \text{clip}\left(r_{t}^{(i,\ell)}(\theta); \epsilon\right)\hat{A}^{(i,\ell)}\right\} - \beta D_{\text{KL}}(\pi_{\theta} || \pi_{\text{ref}})\right)\right].
\end{align}
\normalsize

The advantage $\hat{A}^{(i,\ell)}$ comprises two terms: a term computed jointly across subgroups (i.e., a \textit{global} advantage) and a term computed using only subgroup rewards (i.e., a \textit{local} advantage). These two advantages are \textit{complementary}: The global term ties together all subgroups, serving as the mechanism that affirms state equivalence, whereas the local term ensures that each subgroup's responses are improved independently. That is, the model is taught that all initial states are equivalent and to improve performance for each new initial state. Notably, only using the global advantage may lead to performance degradation, as we show in~\Cref{sec:exp:analysis}. Two key behaviors arise from the above objective:
\begin{itemize}[leftmargin=*,noitemsep,topsep=5pt]
    \item \textbf{Inherent output diversity:} The set of outputs across all subgroups is necessarily more diverse than a set sampled from just the original state. Such diversity was found to help in RL training~\citep{liu2025noisyrollout}.
    \item \textbf{Learned state equivalence:} By comparing outputs from different $\mathcal{T}_\ell(q)$ against each other via $\textcolor{orange}{\bar{R}^{[L]}}$ and $\textcolor{orange}{\sigma_{R^{[L]}}}$, each transformed state is modeled as \textit{equivalent} in terms of outcome. Paired with updating the policy with respect to $\textcolor{blue}{\mathcal{T}_{\ell}(q)}$, the model is trained to be \textit{transformation invariant}.
\end{itemize}
We note that some transformations may change the answer $a$ based on $\mathcal{T}_\ell$, which we denote as $a^{(\ell)}$. As an example, consider applying \method to train a model to be multiple choice order invariant (i.e., not sensitive to if the correct answer is choice $A$, $B$, $C$, or $D$), where each $\mathcal{T}_\ell$ corresponds to a different multiple choice ordering. Concretely, consider the toy example: 
\small
\begin{align*}
    \text{Which animal flies?} \qquad \mathcal{T}_1: \text{[A] Cat; [B] Dog; [C] Bird; [D] Pig} \qquad \mathcal{T}_2: \text{[A] Bird; [B] Pig; [C] Dog; [D] Cat}.
\end{align*}
\normalsize
Here, the correct answer \textit{label} must change with each transformation ($a^{(1)}=[\text{C}]$ and $a^{(2)}=[\text{A}]$). However, the \textit{substance} of the correct answer (``Bird'') is the same across subgroups, making the subgroups comparable and initial states substantively equivalent. 

\textbf{Connections to existing work.} Using data augmentations is inspired by SimCLR~\citep{chen2020simple}, a contrastive learning framework that maximizes representation similarity between different augmentations (e.g., rotations) of the same image. While SimCLR is explicitly contrastive, the contrastive nature of \method is more implicit through the advantage and policy update. \method can be considered a policy gradient algorithm that leverages \textit{state-equivalence}, connecting it to a line of work in exploiting such structure in Markov Decision Processes~\citep{li2006towards,ravindran2004approximate,abel2016near,asadi2019model} and RL~\citep{brunskill2013sample,mandel2016efficient}. Finally, we note that concurrent work, NoisyRollout~\citep{liu2025noisyrollout}, in Vision Language Model RL has explored injecting noise into the image prior to generating group outputs. The key difference between NoisyRollout and \method is that they \textit{do not condition on the transformed initial state during the policy update}, i.e., they use $r_t^{(i)}(\theta)$ rather than $r_t^{(i,\ell)}(\theta)$ (\Cref{eq:esgrpo_ratio}). This difference may appear small but it is crucial, as omitting state-specific conditioning may lead to sub-optimal or incorrect policy updates. For example, when $\mathcal{T}_\ell(q)$ results in a new answer $a^{(\ell)}$, correct behavior is only encouraged when the policy update is with respect to $\mathcal{T}_\ell(q)$. As such, NoisyRollout does not explicitly teach models that augmented initial states are equivalent; rather, as they note, it serves as a group output diversification strategy.

\section{Judge Training with \methodfull.}\label{sec:method:judge}
\method, as presented above, is a general-purpose policy-gradient algorithm. Here, we describe its application for judge training. Concretely, our goal is to train a pairwise judge, which takes input a triplet $(x, y_1, y_2)$ of user input $x$ and model responses $y_1$ and $y_2$, parsed into prompt template $\mathtt{T}(x, A, B)$, and generates a natural language output $o = \{c, j\}$, where $c$ is a chain-of-thought critique and $j$ is the final judgment ($A$ or $B$). The prompt template $\mathtt{T}(x,A,B)$ includes evaluation and formatting instructions; $x$ is the user input, $A$ is the \textit{first} response, and $B$ is the \textit{second} response. In what follows, $(A=y_i, B=y_j)$ means that response $y_i$ is presented to the judge first, and response $y_j$ is presented second. \textit{Without loss of generality, we assume that $y_1$ is the better response.} To train a judge model, we utilize a judgment dataset $\mathcal{D} = \{(x, y_1, y_2)\}$ of pairwise responses. 

\begin{wrapfigure}{r}{0.39\textwidth}
    \vspace{-1\baselineskip}
    \centering
    \includegraphics[width=0.39\textwidth]{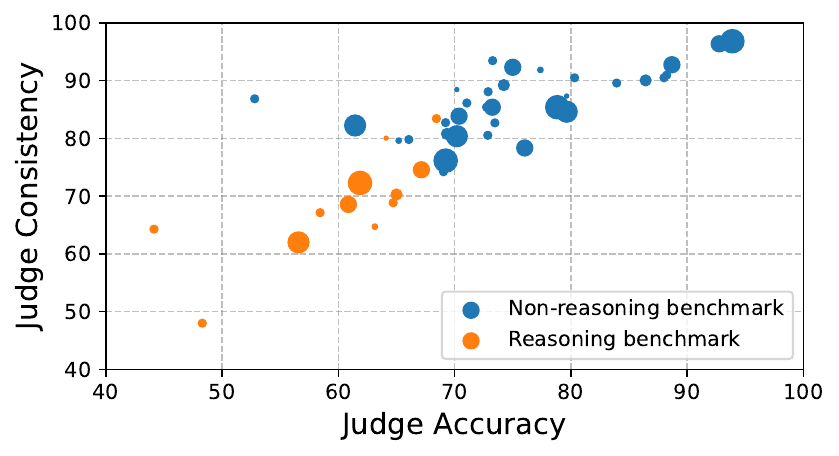}
    \caption{Judge pairwise consistency is inversely correlated with task difficulty, with very poor consistency in reasoning tasks. Plot of 11 judges, with marker size indicating model size across 3 non-reasoning (HHH, LFQA, InstruSum) and 1 reasoning benchmark (JudgeBench).
    }
    \label{fig:consistency}
\end{wrapfigure}
\textbf{When judging reasoning, are judges just guessing?} Past work~\citep{li2023generative,wang2023large} has shown that judge models are sensitive to the order of responses. For each pairwise sample, there are two possible orderings: $(A=y_1, B=y_2)$ and $(A=y_2, B=y_1)$. Merely swapping the order can elicit a different chosen response $y$ from the judge. As a result, past work has measured judge \textit{consistency}: the percentage of pairwise samples the judge chooses the same response in both orderings, with higher consistency indicating less positional bias. In~\Cref{fig:consistency}, we visualize judge performance on 3 non-reasoning benchmarks (HHH~\citep{askell2021general}, LFQA~\citep{xu2023critical}, InstruSum~\citep{liu2023benchmarking}) and a reasoning-focused benchmark, JudgeBench~\citep{tan2024judgebench}. The trend is clear: Reasoning evaluation is uniquely difficult, with judges exhibiting behavior more akin to random guessing than substantive assessment, as measured by consistency.

Judge inconsistency is not a newly discovered phenomenon, with recent judge work taking at proactive measures to mitigate this bias during training, coalescing around using data duplication~\citep{li2023generative,park2024offsetbias,saha2025learning} or simply ensuring balanced datasets at larger data scales~\citep{wang2024direct,cao2024compassjudger}. The former involves including both $(A=y_1, B=y_2)$ and $(A=y_2, B=y_1)$ orderings in the training set as separate samples, doubling the size of the training set, whereas the latter does not explicitly expose swapped order pairs to the judge to save on training time. 
Despite targeted training strategies, judges are still not able to consistently assess as task difficulty grows as shown in \Cref{fig:consistency}.

\textbf{\method for positionally robust and outcome supervised judge training.}
Data-duplication methods suffer from a key drawback: There is no way of explicitly linking together a train sample and its position-swapped counterpart; rather, the model consumes each sample independently during training. \method is uniquely suited to bridge this gap, as it explicitly trains the judge to be order invariant, leading to higher consistency \textit{and} better accuracy, as shown in~\Cref{sec:exp:analysis}. A natural set of transformations $\mathcal{T}$ in judge training is the order response pairs in the input $(x,y_1,y_2)$. 
Recall that the judge is tasked with choosing which of responses $y_1$ or $y_2$ is better, where we assume WLOG that $y_1$ is better. To instill order invariance in the judge, one can use 2 subgroups where subgroup 1 orders responses as $(A = y_1, B = y_2)$, with $a^{(1)} = A$, and subgroup 2 orders responses as $(A = y_2, B = y_1)$, with $a^{(2)} = B$. 

To train \judgeacc{}, we utilize the above subgroup formulation to mitigate inconsistency. We employ both a judgment reward $\mathcal{R}_j$ and format reward $\mathcal{R}_f$:
\begin{align}
    \mathcal{R}_j(o, a) = \begin{cases}
        1, \quad \texttt{parse}(o) = a \\
        0, \quad \texttt{parse}(o) \neq a
    \end{cases}
    \quad \text{and} \quad 
    \mathcal{R}_f(o) = \begin{cases}
        \ \ \ 0.5, \quad o\ \text{formatted correctly}\\
        -0.5, \quad \text{otherwise}
    \end{cases},
\end{align}
where $\texttt{parse$(o)$}$ parses the final judgment from the output $o$. We compute the final reward $R^{(i,\ell)} = \mathcal{R}_j(o^{(i,\ell)}, a^{(\ell)}) + \mathcal{R}_f(o^{(i,\ell)})$ and use $G=32$, with $16$ for each ordering of responses. 

For training data, we construct a simple reasoning-centric dataset: We sample $20$ outputs from Qwen2.5-7B,72B~\citep{yang2024qwen2} and Llama-3.3-70B~\citep{dubey2024llama} using the ReClor~\citep{yu2020reclor} and MATH~\citep{hendrycks2021measuring} \textit{train} sets, forming correct and incorrect pairs based on outcome labels and removing samples with all the same outcome (i.e., all correct or incorrect). This yields a compact training set of $\sim10K$ pairwise samples. We use rollout batch size of 256, train batch size of 128, constant learning rate of $1e-6$, clip $\epsilon=0.2$, and the KL $\beta = 0.0001$. See~\Cref{app:training_details:grpo} for more details. We train two judge variants: \judge{}, which is initialized from CompassJudger-7B~\citep{cao2024compassjudger} and \judgeins{}, which is initialized from Qwen2.5-7B-Instruct~\citep{yang2024qwen2}. 
For both judges, we use a version of the Arena-Hard~\citep{li2024crowdsourced} judge prompt template that omits ties; See~\Cref{app:prompt}. As we show in~\Cref{sec:exp}, \method{} is effective for both continual finetuning and ground-up judge training.

\subsection{Diversifying Reasoning Evaluation with \benchmark{}}\label{sec:benchmark}

Recent benchmarks have focused largely on evaluating judge robustness to stylistic factors (e.g.,~\cite{liu2024rm,zeng2023evaluating}), with reasoning splits focusing on (relatively easy) math and coding benchmarks (e.g., using HumanEvalPack~\citep{muennighoff2023octopack} prompts). JudgeBench~\citep{tan2024judgebench} takes a step in diversifying reasoning tasks covered, including knowledge-based reasoning (via MMLU-Pro~\citep{wang2024mmlu}) and more difficult data sources (e.g., LiveBench~\citep{white2024livebench}). The key insight of JudgeBench is that if a strong model struggles to answer a question, that same question will be difficult to assess for the judges. Based on this insight, they construct JudgeBench in a two-stage manner: (1) Sample multiple outputs from a strong LLM for a question with an objectively correct answer (e.g., math questions) and (2) create pairwise test samples by pairing together a correct and incorrect model response, as determined by final outcome.

While JudgeBench presents a challenge, it comprises only 350 pairwise samples that cover only math, code, knowledge-based QA, and logic puzzles, a subset of the myriad of reasoning tasks available. To diversify evaluation, we use the JudgeBench data creation pipeline and GPT-4o to produce \benchmark{}, which consists of 1,483 challenging response pairs sourced from both established (ARC-Challenge~\citep{clark2018think}, ReClor~\citep{yu2020reclor}, StrategyQA~\citep{geva2021did}, Folio~\citep{han2022folio}) and newer benchmarks (OlympiadBench~\citep{he2024olympiadbench}, AIME 2024 and 2025, SuperGPQA~\citep{du2025supergpqa}, BIG-Bench Extra Hard~\citep{kazemi2025big}). In all, \benchmark{} consists of 4 splits: Math, multi-hop, domain-specific, and ``everyday'' reasoning (inductive, causal, and common-sense); See~\Cref{fig:overview} (center) for a split-level breakdown and~\Cref{app:benchmark} for further details.
\begin{table}[t!]
\caption{ \judge{} is the best performing small judge, almost matching the performance of the best large judges despite being 4.5x smaller in size. Best and second-best in each size range are \textbf{bolded} and \underline{underlined}. \texttt{Avg.} is a microaverage across resp. benchmark splits, \texttt{Overall} is an average of the 3 \texttt{Avg.} scores. Highlighting indicates training method: \hlc[Goldenrod!30]{SFT}, \hlc[orange!50]{DPO}, and \hlc[ForestGreen!20]{RL}. $^\dagger$ indicates that a judge had systematic issues with output format. All values are \textit{consistent accuracy}.}
\label{tab:main_result}

\vspace{0.1in}

\resizebox{\textwidth}{!}{%
\begin{tabular}{llccccccccccccccc}
\toprule
 &
   &
  \multicolumn{4}{c}{\textbf{PPE}} &
  \multicolumn{5}{c}{\textbf{JudgeBench}} &
  \multicolumn{5}{c}{\textbf{ReasoningJudgeBench}} &
  \multicolumn{1}{c}{} \\
\multirow{-2}{*}{} &
  \multirow{-2}{*}{\textbf{Model}} &
  {\color[HTML]{9B9B9B} MBPP} &
  {\color[HTML]{9B9B9B} MATH} &
  {\color[HTML]{9B9B9B} GPQA} &
  \texttt{Avg.} &
  {\color[HTML]{9B9B9B} Knowledge} &
  {\color[HTML]{9B9B9B} Math} &
  {\color[HTML]{9B9B9B} Reasoning} &
  {\color[HTML]{9B9B9B} Code} &
  \texttt{Avg.} &
  {\color[HTML]{9B9B9B} Multi-hop} &
  {\color[HTML]{9B9B9B} Math} &
  {\color[HTML]{9B9B9B} Everyday} &
  {\color[HTML]{9B9B9B} Domain} &
  \texttt{Avg.} &
  \multicolumn{1}{c}{\multirow{-2}{*}{\texttt{Overall}}} \\ \midrule
 \multirow{8}{*}{\rotatebox{90}{\small Small Judges}} &
  \cellcolor{Goldenrod!30}Prometheus-2-7B &
  {\color[HTML]{9B9B9B} 29.70} &
  {\color[HTML]{9B9B9B} 33.01} &
  {\color[HTML]{9B9B9B} 27.54} &
  30.09 &
  {\color[HTML]{9B9B9B} 16.23} &
  {\color[HTML]{9B9B9B} 33.93} &
  {\color[HTML]{9B9B9B} 28.57} &
  {\color[HTML]{9B9B9B} 35.71} &
  24.86 &
  {\color[HTML]{9B9B9B} 16.56} &
  {\color[HTML]{9B9B9B} 29.23} &
  {\color[HTML]{9B9B9B} 17.23} &
  {\color[HTML]{9B9B9B} 16.33} &
  18.88 &
  24.61 \\
 &
  \cellcolor{Goldenrod!30}CompassJudger-7B &
  {\color[HTML]{9B9B9B} 33.89} &
  {\color[HTML]{9B9B9B} 57.27} &
  {\color[HTML]{9B9B9B} 40.86} &
  44.02 &
  {\color[HTML]{9B9B9B} 45.45} &
  {\color[HTML]{9B9B9B} 73.21} &
  {\color[HTML]{9B9B9B} 45.92} &
  {\color[HTML]{9B9B9B} 38.10} &
  49.14 &
  {\color[HTML]{9B9B9B} 34.95} &
  {\color[HTML]{9B9B9B} 50.38} &
  {\color[HTML]{9B9B9B} 32.62} &
  {\color[HTML]{9B9B9B} 38.00} &
  37.76 &
  43.64 \\
 &
  \cellcolor{orange!50}RISE-Judge-7B &
  {\color[HTML]{9B9B9B} 41.18} &
  {\color[HTML]{9B9B9B} 61.21} &
  {\color[HTML]{9B9B9B} 36.56} &
  46.35 &
  {\color[HTML]{9B9B9B} 44.81} &
  {\color[HTML]{9B9B9B} 57.14} &
  {\color[HTML]{9B9B9B} 35.71} &
  {\color[HTML]{9B9B9B} 47.62} &
  44.57 &
  {\color[HTML]{9B9B9B} 28.60} &
  {\color[HTML]{9B9B9B} 51.15} &
  {\color[HTML]{9B9B9B} 31.69} &
  {\color[HTML]{9B9B9B} 36.00} &
  34.73 &
  41.88 \\
 &
  \cellcolor{ForestGreen!20}JudgeLRM-7B$^\dagger$ &
  {\color[HTML]{9B9B9B} 28.80} &
  {\color[HTML]{9B9B9B} 40.47} &
  {\color[HTML]{9B9B9B} 24.77} &
  31.37 &
  {\color[HTML]{9B9B9B} 25.32} &
  {\color[HTML]{9B9B9B} 44.64} &
  {\color[HTML]{9B9B9B} 34.69} &
  {\color[HTML]{9B9B9B} 47.62} &
  33.71 &
  {\color[HTML]{9B9B9B} 5.52} &
  {\color[HTML]{9B9B9B} 20.77} &
  {\color[HTML]{9B9B9B} 14.77} &
  {\color[HTML]{9B9B9B} 13.67} &
  11.87 &
  25.65 \\
 &
  \cellcolor{ForestGreen!20}RM-R1-Distill-7B$^\dagger$ &
  {\color[HTML]{9B9B9B} 31.44} &
  {\color[HTML]{9B9B9B} 55.04} &
  {\color[HTML]{9B9B9B} 15.23} &
  33.96 &
  {\color[HTML]{9B9B9B} 17.53} &
  {\color[HTML]{9B9B9B} 16.07} &
  {\color[HTML]{9B9B9B} 19.39} &
  {\color[HTML]{9B9B9B} 4.76} &
  16.29 &
  {\color[HTML]{9B9B9B} 12.37} &
  {\color[HTML]{9B9B9B} 35.38} &
  {\color[HTML]{9B9B9B} 14.46} &
  {\color[HTML]{9B9B9B} 3.67} &
  15.10 &
  21.78 \\
 &
  \cellcolor{Goldenrod!30}CompassJudger-14B &
  {\color[HTML]{9B9B9B} 36.84} &
  {\color[HTML]{9B9B9B} 59.57} &
  {\color[HTML]{9B9B9B} 40.47} &
  45.64 &
  {\color[HTML]{9B9B9B} 46.75} &
  {\color[HTML]{9B9B9B} 73.21} &
  {\color[HTML]{9B9B9B} 45.92} &
  {\color[HTML]{9B9B9B} 42.86} &
  50.29 &
  {\color[HTML]{9B9B9B} 37.29} &
  {\color[HTML]{9B9B9B} 51.54} &
  {\color[HTML]{9B9B9B} 28.92} &
  {\color[HTML]{9B9B9B} 36.00} &
  37.69 &
  44.54 \\
 &
  \cellcolor{ForestGreen!20}RM-R1-Distill-14B &
  {\color[HTML]{9B9B9B} 32.94} &
  {\color[HTML]{9B9B9B} 72.89} &
  {\color[HTML]{9B9B9B} 37.46} &
  {\underline{47.80}} &
  {\color[HTML]{9B9B9B} 42.21} &
  {\color[HTML]{9B9B9B} 57.14} &
  {\color[HTML]{9B9B9B} 50.00} &
  {\color[HTML]{9B9B9B} 42.86} &
  46.86 &
  {\color[HTML]{9B9B9B} 47.16} &
  {\color[HTML]{9B9B9B} 68.08} &
  {\color[HTML]{9B9B9B} 34.77} &
  {\color[HTML]{9B9B9B} 25.33} &
  {\underline{43.70}} &
  {\underline{46.12}} \\
 \arrayrulecolor{gray}\hdashline
    \arrayrulecolor{black}
  \multirow{2}{*}{\rotatebox{90}{\small {Ours\phantom{1}}}} &
  \cellcolor{ForestGreen!20}J4R-Qwen-Inst-7B (ours) &
  {\color[HTML]{9B9B9B} 29.19} &
  {\color[HTML]{9B9B9B} 60.43} &
  {\color[HTML]{9B9B9B} 43.95} &
  43.95 &
  {\color[HTML]{9B9B9B} 51.30} &
  {\color[HTML]{9B9B9B} 62.50} &
  {\color[HTML]{9B9B9B} 52.04} &
  {\color[HTML]{9B9B9B} 42.86} &
  {\underline{52.29}} &
  {\color[HTML]{9B9B9B} 38.96} &
  {\color[HTML]{9B9B9B} 59.62} &
  {\color[HTML]{9B9B9B} 31.38} &
  {\color[HTML]{9B9B9B} 46.33} &
  41.74 &
  45.99 \\
 &
  \cellcolor{ForestGreen!20}J4R-CJ-7B (Ours) &
  {\color[HTML]{9B9B9B} 37.71} &
  {\color[HTML]{9B9B9B} 64.41} &
  {\color[HTML]{9B9B9B} 45.82} &
  \textbf{49.14} &
  {\color[HTML]{9B9B9B} 55.19} &
  {\color[HTML]{9B9B9B} 71.43} &
  {\color[HTML]{9B9B9B} 54.08} &
  {\color[HTML]{9B9B9B} 56.86} &
  \textbf{56.86} &
  {\color[HTML]{9B9B9B} 39.80} &
  {\color[HTML]{9B9B9B} 61.15} &
  {\color[HTML]{9B9B9B} 36.92} &
  {\color[HTML]{9B9B9B} 42.33} &
  \textbf{45.04} &
  \textbf{50.35} \\ \midrule
 \multirow{6}{*}{\rotatebox{90}{\small Large Judges}} &
  \cellcolor{orange!50}RISE-Judge-32B &
  {\color[HTML]{9B9B9B} 43.63} &
  {\color[HTML]{9B9B9B} 69.96} &
  {\color[HTML]{9B9B9B} 43.93} &
  {\underline{52.53}} &
  {\color[HTML]{9B9B9B} 46.75} &
  {\color[HTML]{9B9B9B} 60.71} &
  {\color[HTML]{9B9B9B} 36.73} &
  {\color[HTML]{9B9B9B} 52.38} &
  46.86 &
  {\color[HTML]{9B9B9B} 45.65} &
  {\color[HTML]{9B9B9B} 55.38} &
  {\color[HTML]{9B9B9B} 27.69} &
  {\color[HTML]{9B9B9B} 40.33} &
  42.35 &
  47.25 \\
 &
  \cellcolor{Goldenrod!30}CompassJudger-32B &
  {\color[HTML]{9B9B9B} 41.10} &
  {\color[HTML]{9B9B9B} 65.16} &
  {\color[HTML]{9B9B9B} 45.66} &
  50.66 &
  {\color[HTML]{9B9B9B} 50.00} &
  {\color[HTML]{9B9B9B} 69.64} &
  {\color[HTML]{9B9B9B} 53.06} &
  {\color[HTML]{9B9B9B} 54.76} &
  \underline{54.57} &
  {\color[HTML]{9B9B9B} 49.16} &
  {\color[HTML]{9B9B9B} 60.77} &
  {\color[HTML]{9B9B9B} 38.15} &
  {\color[HTML]{9B9B9B} 38.00} &
  \textbf{46.53} &
  {\underline{50.59}} \\
 &
  \cellcolor{ForestGreen!20}RM-R1-Distill-32B &
  {\color[HTML]{9B9B9B} 44.10} &
  {\color[HTML]{9B9B9B} 78.95} &
  {\color[HTML]{9B9B9B} 40.94} &
  \textbf{54.71} &
  {\color[HTML]{9B9B9B} 45.45} &
  {\color[HTML]{9B9B9B} 64.29} &
  {\color[HTML]{9B9B9B} 60.20} &
  {\color[HTML]{9B9B9B} 59.52} &
  {54.29} &
  {\color[HTML]{9B9B9B} 52.51} &
  {\color[HTML]{9B9B9B} 69.23} &
  {\color[HTML]{9B9B9B} 33.85} &
  {\color[HTML]{9B9B9B} 28.00} &
  {\underline{46.39}} &
  \textbf{51.80} \\
 &
  \cellcolor{Goldenrod!30}Prometheus-2-8x7B &
  {\color[HTML]{9B9B9B} 27.02} &
  {\color[HTML]{9B9B9B} 37.34} &
  {\color[HTML]{9B9B9B} 31.01} &
  31.79 &
  {\color[HTML]{9B9B9B} 37.01} &
  {\color[HTML]{9B9B9B} 57.14} &
  {\color[HTML]{9B9B9B} 37.76} &
  {\color[HTML]{9B9B9B} 21.43} &
  38.57 &
  {\color[HTML]{9B9B9B} 20.74} &
  {\color[HTML]{9B9B9B} 38.08} &
  {\color[HTML]{9B9B9B} 26.46} &
  {\color[HTML]{9B9B9B} 25.33} &
  25.96 &
  32.11 \\
 &
  \cellcolor{orange!50}Self-Taught-70B &
  {\color[HTML]{9B9B9B} 41.93} &
  {\color[HTML]{9B9B9B} 50.08} &
  {\color[HTML]{9B9B9B} 41.68} &
  44.57 &
  {\color[HTML]{9B9B9B} 50.65} &
  {\color[HTML]{9B9B9B} 58.93} &
  {\color[HTML]{9B9B9B} 42.86} &
  {\color[HTML]{9B9B9B} 40.48} &
  48.57 &
  {\color[HTML]{9B9B9B} 45.32} &
  {\color[HTML]{9B9B9B} 39.62} &
  {\color[HTML]{9B9B9B} 28.31} &
  {\color[HTML]{9B9B9B} 35.67} &
  38.64 &
  43.93 \\
 &
  \cellcolor{orange!50}EvalPlanner-3.3-70B &
  {\color[HTML]{9B9B9B} -} &
  {\color[HTML]{9B9B9B} -} &
  {\color[HTML]{9B9B9B} -} &
  - &
  {\color[HTML]{9B9B9B} 55.8} &
  {\color[HTML]{9B9B9B} 69.6} &
  {\color[HTML]{9B9B9B} 56.1} &
  {\color[HTML]{9B9B9B} 42.9} &
  \textbf{56.6} &
  {\color[HTML]{9B9B9B} -} &
  {\color[HTML]{9B9B9B} -} &
  {\color[HTML]{9B9B9B} -} &
  {\color[HTML]{9B9B9B} -} &
  - &
  - \\ \midrule
 \multirow{6}{*}{\rotatebox{90}{\small Prompted Judges}} &
  \cellcolor[HTML]{EFEFEF}Qwen2.5-7B-Instruct &
  {\color[HTML]{9B9B9B} 32.54} &
  {\color[HTML]{9B9B9B} 48.59} &
  {\color[HTML]{9B9B9B} 34.65} &
  38.61 &
  {\color[HTML]{9B9B9B} 35.06} &
  {\color[HTML]{9B9B9B} 44.64} &
  {\color[HTML]{9B9B9B} 42.86} &
  {\color[HTML]{9B9B9B} 35.71} &
  38.86 &
  {{\color[HTML]{9B9B9B} 29.60}} &
  {{\color[HTML]{9B9B9B} 33.85}} &
  {{\color[HTML]{9B9B9B} 29.85}} &
  {{\color[HTML]{9B9B9B} 28.67}} &
  {30.21} &
  35.89 \\
 &
  \cellcolor[HTML]{EFEFEF}GPT-4o-mini &
  {\color[HTML]{9B9B9B} 39.68} &
  {\color[HTML]{9B9B9B} 55.86} &
  {\color[HTML]{9B9B9B} 41.37} &
  45.66 &
  {\color[HTML]{9B9B9B} 51.30} &
  {\color[HTML]{9B9B9B} 60.71} &
  {\color[HTML]{9B9B9B} 44.90} &
  {\color[HTML]{9B9B9B} 50.00} &
  50.86 &
  {{\color[HTML]{9B9B9B} 51.34}} &
  {{\color[HTML]{9B9B9B} 45.77}} &
  {{\color[HTML]{9B9B9B} 40.00}} &
  {{\color[HTML]{9B9B9B} 36.67}} &
  {40.39} &
  42.24 \\
 &
  \cellcolor[HTML]{EFEFEF}GPT-4o &
  {\color[HTML]{9B9B9B} 44.10} &
  {\color[HTML]{9B9B9B} 61.17} &
  {\color[HTML]{9B9B9B} 47.23} &
  50.86 &
  {\color[HTML]{9B9B9B} 50.65} &
  {\color[HTML]{9B9B9B} 57.14} &
  {\color[HTML]{9B9B9B} 46.94} &
  {\color[HTML]{9B9B9B} 47.62} &
  50.29 &
  {{\color[HTML]{9B9B9B} 40.13}} &
  {{\color[HTML]{9B9B9B} 54.23}} &
  {{\color[HTML]{9B9B9B} 32.31}} &
  {{\color[HTML]{9B9B9B} 39.33}} &
  {45.25} &
  47.18 \\
 &
  \cellcolor[HTML]{EFEFEF}o1 &
  {\color[HTML]{9B9B9B} 52.57} &
  {\color[HTML]{9B9B9B} 94.92} &
  {\color[HTML]{9B9B9B} 71.02} &
  \textbf{72.90} &
  {\color[HTML]{9B9B9B} 78.57} &
  {\color[HTML]{9B9B9B} 92.86} &
  {\color[HTML]{9B9B9B} 82.65} &
  {\color[HTML]{9B9B9B} 92.86} &
  \textbf{83.71} &
  {{\color[HTML]{9B9B9B} 77.59}} &
  {{\color[HTML]{9B9B9B} 90.38}} &
  {{\color[HTML]{9B9B9B} 58.15}} &
  {{\color[HTML]{9B9B9B} 65.33}} &
  \textbf{73.10} &
  \textbf{76.57} \\
 &
  \cellcolor[HTML]{EFEFEF}o3-mini &
  {\color[HTML]{9B9B9B} 51.08} &
  {\color[HTML]{9B9B9B} 94.61} &
  {\color[HTML]{9B9B9B} 66.05} &
  70.64 &
  {\color[HTML]{9B9B9B} 66.23} &
  {\color[HTML]{9B9B9B} 85.71} &
  {\color[HTML]{9B9B9B} 79.59} &
  {\color[HTML]{9B9B9B} 92.86} &
  {\underline{76.28}} &
  {{\color[HTML]{9B9B9B} 65.72}} &
  {{\color[HTML]{9B9B9B} 91.92}} &
  {{\color[HTML]{9B9B9B} 49.85}} &
  {{\color[HTML]{9B9B9B} 57.33}} &
  \underline{65.14} &
  \underline{70.69} \\
 &
  \cellcolor[HTML]{EFEFEF}Deepseek-R1 &
  {\color[HTML]{9B9B9B} 59.96} &
  {\color[HTML]{9B9B9B} 93.52} &
  {\color[HTML]{9B9B9B} 62.07} &
  {\underline{71.89}} &
  {\color[HTML]{9B9B9B} 71.43} &
  {\color[HTML]{9B9B9B} 89.29} &
  {\color[HTML]{9B9B9B} 73.47} &
  {\color[HTML]{9B9B9B} 71.43} &
  74.86 &
  {{\color[HTML]{9B9B9B} 65.05}} &
  {{\color[HTML]{9B9B9B} 86.15}} &
  {{\color[HTML]{9B9B9B} 40.00}} &
  {{\color[HTML]{9B9B9B} 41.67}} &
  58.53 &
  {68.43}\\\bottomrule
\end{tabular}%
}
\vspace{-1\baselineskip}
\end{table}

\section{Experimental Results and Analysis}\label{sec:exp}

\textbf{Baselines.}
We compare \judge{} against multiple strong, contemporary pairwise judge models: Prometheus-7B, 8x7B~\citep{kim2024prometheus}, CompassJudger-7B, 14B, 32B~\citep{cao2024compassjudger}, RISE-Judge-7B, 32B~\citep{yu2025improve}, Self-Taught-Evaluator-70B~\citep{wang2024self}, EvalPlanner~\citep{saha2025learning}, JudgeLRM-7B~\citep{chen2025judgelrm}, and RM-R1-7B, 14B, 32B~\citep{chen2025rmr1rewardmodelingreasoning}. We report official numbers when they exist, or run each judge with its associated prompt template otherwise\footnote{
    Because our judge is initialized with CompassJudger, we re-run CompassJudger with our prompt, resulting in different results for JudgeBench. Our prompt yields a 3.14 point gain for the 7B model and 4.00 and 4.57 point drops for 14B and 32B from our reproduced results. Extended discussion in~\Cref{app:prompt:tj}.
}. For our main results, we also evaluate o1, o3-mini, GPT-4o, 4o-mini, Qwen2.5-7B~\citep{yang2024qwen2}, and DeepSeek-R1~\citep{guo2025deepseek} on all benchmarks, using our prompt (\Cref{app:prompt}). We additionally evaluate a suite of models on \benchmark{}, which we present in~\Cref{app:benchmark:additional_results}. Unless otherwise specified, all general-purpose models are instruction tuned versions.

\textbf{Benchmarks.}
To assess judge performance in reasoning domains, we use 3 benchmarks: PPE Best-of-$K$~\citep{frick2024evaluate} (MBPP, MATH, and GPQA splits), JudgeBench~\citep{tan2024judgebench}, and \benchmark{}. PPE Best-of-$K$ contains $\sim$2,500 pairwise samples per split, with judges assessing responses sampled from multiple \textit{weaker} generators, like Llama-3-8B. In contrast, JudgeBench and \benchmark{} require judges to assess responses from a stronger model (GPT-4o). For all benchmarks, we adopt a consistent accuracy setup: Each pairwise sample is evaluated twice, with the order of responses swapped on the second run. Then, a judge is only considered correct if both orders are evaluated correctly. \textit{This setup is the default for JudgeBench, but differs from the original PPE setup, which fixes an ordering; as such, our results for PPE differ from their reported results.} For judges that allow ties, we follow~\cite{tan2024judgebench} and consider a run with a tie to be correct if the non-tie response is correct.

\subsection{How well does \judge{} perform in reasoning settings?}\label{sec:exp:main_result}
Our main results are presented in \Cref{tab:main_result}. \judge{} outperforms all small ($\leq14$B) judge models in aggregate (50.35 vs. the next best 46.12) and roughly matches the performance of the best \textit{large} judge models. In particular, despite being 4.5x smaller in size, \judge{} (50.35) almost matches RM-R1-32B (51.80), which was trained with both warm-up SFT with DeepSeek-R1 distilled reasoning traces and GRPO. Such small gap shows that \textit{RL-training alone is insufficient for ensuring robust reasoning evaluation.} As we show later, when given equal compute, \judge{} is able to greatly surpass all larger models. \judgeins{} (45.99), initialized from a much weaker checkpoint, is the second best performing small judge behind \judge{}, surpassing GPT-4o-mini (42.24) and nearly matching the twice as large RM-R1-14B (46.12). 

\benchmark{} is a significant challenge for judge models, with many capable judges failing to crack 45\%. Judges are relatively strong in assessing math, but struggle on the other three splits, showing that evaluating complex \textit{natural language} reasoning (as opposed to math) is still an open problem. We present evaluations and analysis for many additional models (e.g., GPT-4.1 series, Llama-3.1-405B, DeepSeek-V3) in~\Cref{app:benchmark:additional_results}\austin{TODO}.

\subsection{Additional analysis.}\label{sec:exp:analysis}
\begin{table}[t]
    \centering
    \caption{Comparison on JudgeBench and ReasoningJudgeBench of \method against (top) various common judge training algorithms and (bottom) data-driven ways to improve consistency. Both results showcase the benefits of training with \method. \yilun{We can fit more results in this table. Should we include ReasoningJudgeBench as well here (and even JETTS or others)?}\austin{Yes, will update this when results are ready} \austin{TODO: Fill in - as they finish}}
    
    \vspace{0.1in}
    
    \resizebox{0.7\textwidth}{!}{    
        \begin{tabular}{l c c c c c c}
        \toprule
        \multirow{2}{*}{Training algorithm} &  & & \multicolumn{2}{c}{JudgeBench} & \multicolumn{2}{c}{Reas.JudgeBench} \\
         & & & Acc. & Cons. & Acc. & Cons. \\
        \midrule
        CompassJudger-7B (Initial) & & & 49.14 & 68.86 & 37.69 & 68.58 \\
        +SFT & & & 42.86 & 62.86 & 32.97 & 57.11 \\
        +DPO & & & 46.57 & 68.29 & 38.84 & 69.66 \\
        \midrule
        \multirow{2}{*}{Training data strategy} & Norm. & Norm. & Acc. & Cons. & Acc. & Cons. \\
        & data. size & time & & & & \\
        \midrule
        CompassJudger-7B (Initial) & - & - & 49.14 & 68.86 & 37.69 & 68.58 \\
        +GRPO w/ \texttt{[Bal.]}, $G=32$ & 1x & 1x & 52.00 & 73.43 & 41.34 & 75.52 \\
        +GRPO w/ \texttt{[Dup.]}, $G=16$ & 2x & 1x & 51.14 & 76.00 & 38.84 & 71.61 \\
        +GRPO w/ \texttt{[Dup.]}, $G=32$ & 2x & 2x & 51.71 & 77.71 & 42.01 & 77.95 \\
        +GRPO w/ global adv., $G=32$ & 1x & 1x & 49.14 & 69.43 & 37.96 & 68.58 \\\midrule
        +\method, $G=32$ (ours) & 1x & 1x & \textbf{56.86} & \textbf{81.14} & \textbf{45.04} & \textbf{80.51} \\
        \bottomrule
        \end{tabular}
    }
    \label{tab:training_recipe}
\vspace{-1\baselineskip}
\end{table}

\begin{table}[t!]
\centering
\label{tab:inf-scaling}
\caption{FLOP-matched evaluation on JudgeBench comparing \judge{} to larger judges that may produce longer outputs. After matching the amount of compute (last column), \judge{} outperforms all larger baselines by even more significant margins.
} 
\vspace{0.1in}
\resizebox{0.7\columnwidth}{!}{%
        \begin{tabular}{lrrrrc}
        \toprule
        Baseline Model & Size & Avg. tokens & $\delta_{\text{inf}}$ & Baseline & J4R Acc. \\
        &  ($M$) & ($T_{\text{inf}}$) & &  Acc. & (FLOP-matched) \\ \midrule
        J4R-CJ & 7B & 270.28 & 1.0 & 56.86 & - \\
        CompassJudger & 14B & 213.48 & 2.0 & 50.29 & 57.87{\color[HTML]{9B9B9B}$\pm$0.013}  (\color{OliveGreen}{$\uparrow$ \phantom{1}7.58})\\
        CompassJudger & 32B & 212.87 & 4.0 & 54.57 & 62.00{\color[HTML]{9B9B9B}$\pm$0.012}  (\color{OliveGreen}{$\uparrow$ \phantom{1}7.43})\\
        RISE-Judge & 32B & 532.62 & 9.0 & 46.86 & 64.10{\color[HTML]{9B9B9B}$\pm$0.009} (\color{OliveGreen}{$\uparrow$ 17.24})\\
        RM-R1 & 14B & 1621.83 & 12.0 & 46.86 & 64.48{\color[HTML]{9B9B9B}$\pm$0.013} (\color{OliveGreen}{$\uparrow$ 17.62})\\
        Self-Taught & 70B & 394.55 & 15.0 & 48.57 & 64.59{\color[HTML]{9B9B9B}$\pm$0.007} (\color{OliveGreen}{$\uparrow$ 16.02})\\
        RM-R1 & 32B & 1792.37 & 30.0 & 54.29 & 64.94{\color[HTML]{9B9B9B}$\pm$0.005} (\color{OliveGreen}{$\uparrow$ 10.65})\\ \bottomrule
        \end{tabular}%
}
\vspace{-1\baselineskip}
\end{table}

\textbf{How does initial model impact performance?} 
As discussed in~\Cref{sec:method:judge}, we start training from CompassJudger-7B for \judge{} and Qwen2.5-7B for \judgeins{}. From~\Cref{tab:main_result}, we find that starting from a stronger judge checkpoint yields results in better aggregate performance (50.35 vs. 45.99). This result is perhaps not surprising given recent analysis~\citep{yue2025does}, which suggests that a primary benefit of RLVR is modifying the model output distribution to favor correct reasoning chains that are already within the initial model's capabilities.
Hence, initializing RL training from a trained judge provides a ``higher floor'' from which RL can incentivize correct judgments. However, this does not mean \method{} fails with weaker initial models: \judgeins{} improves Qwen2.5-7B from 35.89 aggregate accuracy to 45.99, a 28\% gain, and \textit{outperforms all other $7B$ judges}, some trained with 20x more data. In all, these results show that past work in training judges is not in vain. Existing judge models that perform well in non-reasoning settings can be continually trained with \method{} to become capable reasoning evaluators. We show next that \method{} is \textit{unique} in its ability to ``unlock'' the reasoning evaluation abilities of existing judges: Existing judge training recipes actually cause \textit{degradation} with the same seed dataset.

\textbf{How does \method{} compare with existing judge training approaches?} 
Here, we quantify the advantage \method{} brings over two common judge training recipes, distillation-based SFT and DPO. Prior work in training judges distills CoT judgments from a teacher model, then uses either SFT (e.g.,~\cite{kim2023prometheus,cao2024compassjudger,vu2024foundational}) or DPO (e.g.,~\cite{wang2024direct,hu2024themis,ye2024beyond}). Using our training set, we follow the approach of~\cite{wang2024direct,hu2024themis,ye2024beyond} and distill CoT judgments from Llama-3.1-70B to form both a SFT training set and a paired DPO training set. We ensure each of these training sets are label-balanced, and train for 1 epoch starting from CompassJudger-7B;~\Cref{app:training_details:dpo_sft} contains complete details. 
\Cref{tab:training_recipe} (top) shows the fundamental limitations of current judge model paradigms: Even using targeted reasoning data, SFT and DPO actually \textit{decrease} performance on JudgeBench relative to the baseline performance. This perhaps surprising result likely stems from two factors: teacher-model distillation (1) is \textit{out-of-distribution} finetuning and (2) places an inherent upper-bound on performance. 
Given that many popular instruction-tuned models are relatively weak reasoning evaluators (See~\Cref{app:benchmark:additional_results}), this result highlights limitations of current distillation-based judge training approaches.

\textbf{What benefits does \method bring over other data-driven bias-mitigation strategies?} 
As discussed in~\Cref{sec:method:judge}, previous work has used label-balanced training~\cite{wang2024direct,cao2024compassjudger} or data duplication~\cite{li2023generative,park2024offsetbias,saha2025learning} to mitigate inconsistency. Here, we show that \method improves upon both of these approaches for RL. From our train set, we form a data balanced version (denoted \texttt{[Bal]}), where we assign 50\% of our dataset to have label A and 50\% label B, and (2) a duplicated variant (denoted \texttt{[Dup]}), which doubles the size of our training set by including each ordering of responses. For \texttt{[Bal]}, we run GRPO with the same training setup as \judge{}, as described in~\Cref{sec:method:judge}. For \texttt{[Dup]}, we consider two setups. The first uses $G = 16$ and doubles the rollout batch size, resulting in the same number of model updates and training steps as \judge{}. The second applies the exact same training setup as \judge{}, resulting in twice the training steps. 

We report normalized training time with respect to \judge{}, along with judge performance, in~\Cref{tab:training_recipe} (bottom). Both the balanced dataset and data duplication approaches result in some gains in accuracy and consistency, but fall significantly short of \method{}. For the same number of model updates, \method{} improves JudgeBench accuracy 9.4\% and 11.2\% over \texttt{[Bal]} and \texttt{[Dup]}, respectively. Further, \method{} teaches the model to be more robust positional bias via learned state equivalence, as highlighted by consistency gains. We additionally exploring using only the global advantage computed across subgroups (i.e., only the first term of $\hat{A}^{(i,\ell)}$ in~\Cref{eq:esgrpo_ratio}), and find that its performance lags all other strategies, highlighting the necessity of including both global and local information in the advantage. We present a deeper exploration for why this failure occurs in~\Cref{app:exp:global_only}.

\textbf{Given equal inference-time compute (FLOPs), how does \judge{} match up with larger judges?} 
Here, we present a FLOP-matched evaluation on JudgeBench. We use the commonly adopted FLOP approximation of $2MT_{\text{inf}}$ \citep{snell2024scaling,sardana2023beyond}, where $M$ is the model size and $T_{\text{inf}}$ is the number of inference tokens. For \judge{} and all judges of size $\geq14B$, we compute the average number of output tokens per sample on JudgeBench, allowing us to compute a FLOP-scaling ratio $\delta_{\text{inf}} = \frac{M_{b}T_{b,\text{inf}}}{M_{\text{J4R}}T_{\text{J4R,inf}}}$, where subscript $b$ denotes baseline and J4R denotes \judge{}. We then sample $\delta_{\text{inf}}$ parallel responses from \judge{} with temperature 1.0 and aggregate via majority vote (i.e., self-consistency~\citep{wang2022self}). To reduce compute time, we sample a fixed set of 128 responses from the \judge{} and randomly subsample for each $\delta_{\text{inf}}$. We repeat each subsampling $100$ times and report mean and standard deviation. \Cref{tab:inf-scaling} shows the benefit of matched compute: the performance of \judge{} improves up to 14\%. When afforded the same compute as RM-R1-32B, it achieves an absolute score of 64.94, a relative improvement of 19.6\%.

\begin{wraptable}{r}{0.47\textwidth}
\vspace{-1\baselineskip}
\centering
\caption{Reasoning-specific training generally improves weaker initial models, but degrades chat evaluation for stronger initial models.}
\label{tab:ood}
\resizebox{0.47\columnwidth}{!}{
\begin{tabular}{lccc}
    \toprule
    Model & HHH & LFQA & InstruSum \\\midrule
    CompassJudger-7B & 83.26 & 70.00 & 64.23 \\
    \judge{} & 77.83 & 67.69 & 66.91\\\midrule
    Qwen2.5-7B-Instruct & 72.40 & 50.77 & 51.82\\
    \judgeins{} & 76.47 & 62.69 & 60.83\\\bottomrule
    \end{tabular}
}
\vspace{0.5\baselineskip}
\caption{Group size $G$ ablation.}
\label{tab:ablation}
\resizebox{0.33\columnwidth}{!}{
\begin{tabular}{rrrrrr}
    \toprule
    $G$ & 16 & 24 & 32 & 64 \\\midrule
    Acc. & 40.93 & 40.66 & 45.04 & 44.71 \\
    Cons. & 74.92 & 72.49 & 80.51 & 79.97 \\
    \bottomrule
    \end{tabular}
}
\vspace{-1\baselineskip}
\end{wraptable}

\textbf{How does reasoning-specific training affect other domains?}
We measure the difference between \judgeacc{} and initial models on LFQA~\citep{xu2023critical} (long-form answers to open-ended questions), HHH~\citep{askell2021general} (safety and helpfulness), and InstruSum~\citep{liu2023benchmarking} (summarization quality). Our results, shown in~\Cref{tab:ood}, are in-line with expectations: Specializing a strong model towards a specific domain will reduce general abilities, as shown with degradations in chat-focused settings (HHH and LFQA), but improves performance when the task is reasoning-demanding, like contextual comprehension. However, when the initial model is weak, ``low-hanging fruit'' gains are available, and reasoning specific training holistically improves performance.

\textbf{How does group size affect \method?}
In~\Cref{tab:ablation}, we use \benchmark{} to quantify the effect of group size $G$. We find that performance increases as $G$ increases, with the largest gain occurring from $24$ and $32$. This highlights the importance of using sufficient $G$ to ensure strong contrastive signals between subgroups. Using a larger group size of $64$ does not yield further gains.

\section{Conclusion} We demonstrate the effectiveness of our proposed \method{} algorithm for training judge model to be robust to positional biases, which improves evaluation ability in reasoning-intensive domains. We also introduce \benchmark{}, a challenging benchmark of 1,483 challenging pairwise samples that diversifies existing reasoning benchmarks. \benchmark{} reveals that judge models have significant room for growth in evaluating in reasoning settings.

\newpage
\bibliographystyle{unsrtnat}
\bibliography{references}

\newpage
\newpage
\appendix

\section{Extended Related Work and Discussion}
\subsection{Related work: Pointwise generative verification.} While pairwise evaluation forms the focus of our work, evaluation can also be done in a pointwise manner. Pointwise generative verifiers are another subset of the generative judge paradigm that assess one response at a time. Earlier work~\citep{kim2024prometheus} explored using per-instance references and rubrics to stabilize evaluation, while more recent work~\citep{hu2024themis} explores larger-scaled training with preference optimization. Pointwise evaluators do not necessarily need to be standalone models: Many pairwise judges (e.g.,~\cite{wang2024direct,vu2024foundational,hu2024themis,kim2024prometheus,li2023generative}) are also capable pointwise evaluators, assigning scores of 1-5 or binary pass/fail scores based on certain evaluation criteria. Past work~\citep{park2024offsetbias} has also found that pointwise training can improve pairwise performance. Concurrently, long CoT distillation has been used to improve such models~\citep{khalifa2025process} for math evaluation. Closely related are critic models, which are models trained at providing feedback to models. Such models are typically trained with human critiques~\citep{wang2023shepherd}, distillation~\citep{xi2024enhancing} or more recently, RL~\citep{mcaleese2024llm,akyurek2023rl4f,yao2023retroformer,xie2025teaching}. 

The efficient pointwise nature makes such evaluators attractive choices in settings like inference-time verification. However, recent work has highlighted the difficult nature of single-instance verification in reasoning intensive domains~\citep{zhou2025evaluating}. The pairwise paradigm offers stability in assessment, as relative comparisons a valuable ``other item'' to anchor assessment against. Broadly, relative (ordinal) assessment has been shown to be easier for humans that pointwise (cardinal) assessment~\citep{shah2016estimation}, a fact that also appears to carry over to generative verifiers.

\subsection{Discussion and Practical Considerations}
Our work demonstrates the effectiveness of \method{} for training positionally robust LLM-as-judge models. As our work is focused on quantifying the performance of our methodology, (analyzing model initialization, quantifying advantages over other training approaches, etc.), we did not optimize our training dataset, opting for a simple approach for generating training pairs used in past work~\citep{saha2025learning}. We anticipate that future data-driven investigations, like that in~\citep{chen2025rmr1rewardmodelingreasoning}, will yield even larger gains. 

Finally, we note that \method{} is naturally impacted by choice of group size $G$, as shown in our last experiment. \method{} takes inspiration from contrastive learning frameworks enforce similarity and dissimilarity within \textit{batches} of samples~\citep{chen2020simple,he2020momentum,lee2019latent}; The group size $G$ in \method{} can be thought of as analogous to contrastive learning batch size. A common criticism of such contrastive learning approaches is that they require large training batch sizes~\citep{gao2021scaling,chen2022we}, necessitating large amounts of GPU VRAM. While \method{} requires relatively large $G$, a key difference between \method{} and contrastive learning approaches is that the \textit{minimum required GPU VRAM does not increase as $G$ increases}. This is because policy updates occur sequentially. Complications with gradient accumulation for contrastive learning approaches, described in~\citep{gao2021scaling}, are absent in \method{}. That is, sampling higher $G$ only requires more \textit{training time}, not increased hardware minimums.

\begin{figure}
    \centering
    \includegraphics[width=0.75\linewidth]{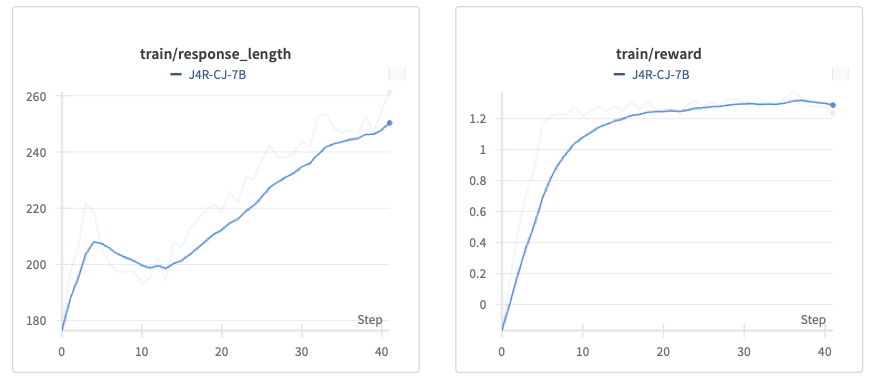}
    \caption{Reward and length dynamics of training \judge{}.}
    \label{fig:training_dynamics}
\end{figure}

\section{Model training details}
\subsection{\method training details}\label{app:training_details:grpo}
To train our judge to be strong in reasoning domains, we create a reasoning-based training set of pairwise model responses. In particular, we sample responses from both weaker (Qwen2.5-7B) and stronger models (Qwen2.5-72B and Llama3.3-70B) to ensure diversity. We choose two datasets with training sets, MATH~\citep{hendrycks2021measuring} and ReClor~\citep{yu2020reclor} with objectively correct answers. We then prompt each model to answer each question in the training set after outputting a CoT, sampling $20$ responses using temperature of $1.0$. Based on final outputs, we form pairs of ``correct'' and ``incorrect'' responses. This can be viewed as a form of synthetic data generation, as employed by prior work, e.g.,~\citep{wang2024self,saha2025learning}. All training was conducted on an 8xH200 GPU node.

\subsection{\method training dymanics}
We visualize training dynamics~\Cref{fig:training_dynamics}, we present the judge response length and average reward dynamics during training. Response length grows steadily as training progresses, with the final judge response length nearly 50\% longer than the starting point. Reward grows steadily over the first quarter of training before transitioning to a more gradual increase. We additionally find that JudgeBench and \benchmark{} performance, both accuracy and consistency, both improve relatively steadily over the course of training.

\subsection{DPO and SFT judge baseline details}\label{app:training_details:dpo_sft}
From our train set, we distill $20$ CoT judgments from Llama-3.1-70B-Instruct per prompt using a temperature of $0.7$ and divide the $20$ teacher outputs into correct and incorrect outputs based on ground-truth label. We then form DPO pairs by selecting one correct and incorrect response. We ensure that the dataset has balanced labels. For samples where the teacher model produces the same judgment for all $20$ samples (i.e., all correct or all incorrect), we include the sample as a non-CoT training sample: We remove the requirement for CoT judgment in the prompt, asking the judge to respond directly with "A" or "B" without CoT. This type of data sample was found to help judge training~\citep{wang2024direct} and allows us to use all of our training data, for fair comparison. We additionally add an SFT loss during DPO, as is commonly done in judge training~\citep{wang2024direct,ye2024beyond}. For DPO, we use $\beta=0.1$. For SFT, we use the correct samples from the DPO train set. We train both SFT and DPO models for 1 epoch using TRL~\citep{von_Werra_TRL_Transformer_Reinforcement} with commonly found hyperparameters (e.g.,~\cite{li2023generative,ivison2023camels}). We use an effective batch size of 64 and utilize AdamW and a cosine learning rate scheduler with peak learning rates of $1e-5$ (SFT) and $5e-7$ (DPO).

\section{\benchmark{}: Details and additional results.}\label{app:benchmark}
Constructing \benchmark{} follows a three stage approach: (1) We select a challenging benchmark with objectively correct outcomes, (2) We prompt a strong LLM (GPT-4o) to provide $20$ CoT responses for each input, (3) We construct pairs using outcome verification, keeping one correct and one incorrect response per prompt. We samples prompts where the model gets all $20$ responses correct or incorrect. Because judges are known to exhibit length bias~\citep{zeng2023evaluating,park2024offsetbias}, when constructing our pairs, we keep the pair of positive and negative responses with \textit{minimal difference in length} in order to create more challenging pairs.

We chose both established and recent benchmarks as our source datasets. This allows us to balance challenging, newer samples with existing samples that are fundamentally challenging, i.e., are still difficult to answer for models despite being publicly released for some time. In particular, we use
\begin{itemize}[leftmargin=*,noitemsep,topsep=5pt]
    \item \textbf{ARC-Challenge}~\citep{clark2018think}: An established multiple-choice QA benchmark.
    \item \textbf{ReClor}~\citep{yu2020reclor} (Validation set): A multiple-choice QA benchmark focused on reading comprehension, sourced from professional standardized tests. 
    \item \textbf{StrategyQA}~\citep{geva2021did}: An implicit, multi-hop reasoning QA benchmark.
    \item \textbf{Folio}~\citep{han2022folio} (Validation set): A human-written first-order logic benchmark
    \item \textbf{AIME 2024 and 2025}: Competition-level math problems
    \item \textbf{OlympiadBench}~\citep{he2024olympiadbench} (Text-only): Olympiad-level math problems.
    \item \textbf{SuperGPQA}~\citep{du2025supergpqa} (Random subsample): An expanded version of GPQA~\citep{rein2024gpqa}, scaled to 285 disciplines.
    \item \textbf{BIG-Bench Extra Hard}~\citep{kazemi2025big} (Boolean Expressions, Causal Understanding~\citep{nie2024moca,kiciman2023causal}, Disambiguation QA, Hyperbaton, Multi-step arithmetic): A harder variant of BigBench-Hard~\citep{suzgun2022challenging}.
\end{itemize}
The above selection of benchmarks was chosen to fill the gaps of evaluation. Namely, JudgeBench~\citep{tan2024judgebench} draws from a small number of benchmarks, limiting the types of reasoning covered. \benchmark{} represents an expansion of reasoning domains evaluated, including the myriad of unique reasoning tasks from BIG-Bench Extra Hard. Due to the limited size and difficulty of AIME 2024 and 2025, GPT-4o was only able to generate 15 valid pairs. As a result, we utilized GPT-4.1-mini to augment math subset. Both models' responses are included in the Math subset. The per-split data source breakdown is as follows, where we abbreviate BIG-Bench Extra Hard as BBEH.
\begin{itemize}[leftmargin=*,noitemsep,topsep=5pt]
    \item \textbf{Multi-hop reasoning} [598 samples]: ARC-Challenge (60), BBEH Boolean Expressions (161), Folio (88), ReClor (289)
    \item \textbf{Math} [260 samples]: AIME 2024 and 2025 (38), BBEH Multi-step arithmetic (38), OlympiadBench (184).
    \item \textbf{Domain} [300 samples]: SuperGPQA (300)
    \item \textbf{Everyday} [325]: BBEH Causal Understanding (89), BBEH Disambiguation QA (68), BBEH Hyperbaton (114), StrategyQA (54).
\end{itemize}

\subsection{Model inference details.} All models of size 72B or smaller were run locally on vLLM~\citep{kwon2023efficient} on nodes of 1,4, or 8xA100s. OpenAI model inference was run using their Batch API with the following endpoints: \texttt{gpt-4o-2024-08-06}, \texttt{gpt-4o-mini-2025-04-16}, \texttt{o3-mini-2025-01-31}, \texttt{o1-2024-12-17}, 
and \texttt{o4-mini-2025-04-16}. All reasoning models are evaluated with ``reasoning effort'' set to medium (default). DeepSeek-R1 and V3 and Llama-3.1-405B were accessed through \url{together.ai}.

\subsection{Additional evaluation results}\label{app:benchmark:additional_results}
Here, we evaluate a suite of instruct and reasoning models on \benchmark{}. We present results in~\Cref{tab:benchmark_full}. We find that reasoning models tend to perform well on evaluation, with top performers being o1 at 73.10\% and o4-mini at 69.39\%. The top models are also the most consistent, most exceeding 80\%. 
Among instruction-tuned judges, GPT-4.1 performs well, passing the performance of DeepSeek-R1. 

\begin{table}[t!]
\caption{Complete evaluation results for ReasoningJudgeBench, covering small and large judges (presented in main body), prompted instruction tuned models, and prompted reasoning models. $\dagger$ denotes systematic issues with output formatting.}
\label{tab:benchmark_full}
\centering
\resizebox{\textwidth}{!}{%
\begin{tabular}{llcccccc}
 & \textbf{Model}                & Multi-hop & Math  & Everyday & Domain & Avg.  & Consistency \\ \midrule
 \multirow{8}{*}{\rotatebox{90}{\small Small Judges}} & Prometheus-2-7B               & 16.56     & 29.23 & 17.23  & 16.33    & 18.88 & 47.61       \\
 & CompassJudger-7B              & 34.95     & 50.38 & 32.62  & 38.00    & 37.76 & 67.16       \\
 & RISE-Judge-7B                 & 28.60     & 51.15 & 31.69  & 36.00    & 34.73 & 66.82       \\
 & JudgeLRM-7B$^\dagger$         & 5.52      & 20.77 & 14.77  & 13.67    & 11.87 & 54.28       \\
 & RM-R1-Distill-7B$^\dagger$    & 12.37     & 35.38 & 14.46  & 3.67     & 15.10 & 48.11       \\
 & CompassJudger-14B             & 37.29     & 51.54 & 28.92  & 36.00    & 37.69 & 63.79       \\
 & RM-R1-Distill-14B             & 47.16     & 68.08 & 34.77  & 25.33    & 43.70 & 65.54       \\
 & \judgeins{} (Ours) & 41.97     & 59.62 & 31.38  & 36.33    & 41.60 & 73.57       \\
 & \judge{} (Ours)       & 43.81     & 61.15 & 36.92  & 42.33    & 45.04 & 80.98       \\ \midrule
 \multirow{5}{*}{\rotatebox{90}{\small Large Judges}} & RISE-Judge-32B                & 45.65     & 55.38 & 27.69  & 40.33    & 42.35 & 73.23       \\
 & CompassJudger-32B             & 49.16     & 60.77 & 38.15  & 38.00    & 46.53 & 74.04       \\
 & RM-R1-Distill-32B             & 52.51     & 69.23 & 33.85  & 28.00    & 46.39 & 66.35       \\
 & Prometheus-2-8x7B             & 20.74     & 38.08 & 26.46  & 25.33    & 25.96 & 54.89       \\
 & Self-Taught-70B               & 45.32     & 39.62 & 28.31  & 35.67    & 38.64 & 63.32       \\ \midrule
 \multirow{13}{*}{\rotatebox{90}{\small Prompted Inst. Judges}} & Qwen2.5-7B-Instruct           & 29.60     & 33.85 & 29.85  & 28.67    & 30.21 & 57.79       \\
 & Qwen2.5-14B-Instruct          & 41.30     & 39.23 & 31.69  & 25.67    & 35.67 & 62.58       \\
 & Qwen2.5-32B-Instruct          & 48.49     & 38.46 & 35.69  & 40.33    & 42.28 & 68.64       \\
 & Qwen2.5-72B-Instruct          & 50.33     & 48.46 & 32.31  & 39.00    & 43.76 & 72.76       \\
 & Llama-3.1-8B-Instruct         & 25.75     & 33.85 & 20.00  & 17.67    & 24.28 & 48.89       \\
 & Llama-3.1-70B-Instruct        & 48.66     & 38.08 & 28.92  & 30.67    & 38.84 & 67.16       \\
 & Llama-3.3-70B-Instruct        & 51.51     & 45.00 & 28.31  & 10.00    & 36.88 & 72.15       \\
 & Llama-3.1-405B-Instruct       & 57.53     & 55.00 & 32.00  & 36.67    & 47.27 & 73.50       \\
 & GPT-4o-mini                   & 40.13     & 45.77 & 40.00  & 36.67    & 40.39 & 70.47       \\
 & GPT-4o                        & 51.34     & 54.23 & 32.31  & 39.33    & 45.25 & 73.50       \\
 & GPT-4.1-nano                  & 26.25     & 42.31 & 32.00  & 27.00    & 30.48 & 50.51       \\
 & GPT-4.1-mini                  & 59.03     & 77.69 & 42.15  & 49.00    & 56.57 & 80.31       \\
 & GPT-4.1                       & 68.39     & 76.54 & 42.15  & 46.67    & 59.68 & 81.73       \\
 & Deepseek-V3                   & 55.52     & 58.85 & 36.92  & 42.00    & 49.29 & 74.51       \\ \midrule
 \multirow{4}{*}{\rotatebox{90}{\parbox{1.25cm}{\small Prompted Reas.\\Judges}}} & o1                            & 77.59     & 90.38 & 58.15  & 65.33    & 73.10 & 87.53       \\
 & o3-mini                       & 65.72     & 91.92 & 49.85  & 57.33    & 65.14 & 83.41       \\
 & o4-mini                       & 74.92     & 89.62 & 52.31  & 59.33    & 69.39 & 82.87       \\
 & Deepseek-R1                   & 65.05     & 86.15 & 40.00  & 41.67    & 58.53 & 72.89       \\ \bottomrule
\end{tabular}%
}
\end{table}

\section{Additional analysis and qualitative examples}\label{app:exp}
\subsection{Why does only using a global advantage fail?}\label{app:exp:global_only}
In~\Cref{sec:exp:analysis}, we showed that running \method{} \textit{without} individual advantages resulted in dramatically different results. Concretely, if one retains the subgroup sampling but calculates the advantage as $\tilde{A}^{(i,\ell)}$, where 
\small
\begin{align}
    \tilde{A}^{(i,\ell)} =  (R^{(i,\ell)} - \textcolor{orange}{\bar{R}^{[L]}}) / \textcolor{orange}{\sigma_{R^{[L]}}},
\end{align}
\normalsize
the performance drops significantly (e.g., from 56.86 on JudgeBench to 49.14). Here we provide an illustrative example into the drawbacks of only using the global advantage. 

Suppose we have two subgroups of 16 responses, denoted subgroup A and subgroup B, where subgroup A's performance is very strong (12 responses of reward 1.5 and 4 responses of reward 1.0) and subgroup B's performance is very poor (12 responses of reward 0.0 and 4 responses of reward 1.0). Now, consider the three candidates for advantage computation: (1) Separate groups (vanilla GRPO), (2) Global advantage only, and (3) Global and Local advantage (\method{}).

\textbf{Separate groups.} Within subgroup A, the reward 1.5 responses will receive advantage 0.577, while the reward 1.0 responses receive advantage -1.732. For subgroup B, the reward 1.0 responses will receive advantage 1.732, while the reward 0.0 responses receive advantage -0.577. Each subgroup behaves as expected: Better than average rewards are encouraged, and worse than average rewards are discouraged.

\textbf{Global advantage only.} Computing a mean and standard deviation across both subgroups yields the following. Within subgroup A, the reward 1.5 responses will receive advantage 1.044, while the reward 1.0 responses receive advantage 0.285. For subgroup B, the reward 1.0 responses will receive advantage 1.732, while the reward 0.0 responses receive advantage -1.234. At first glance, such advantage assignment is intuitive: The reward 1.5 responses get encouraged heavily, while the reward 0.0 responses get heavily discouraged. 

However, the key drawback is that the poor responses in subgroup B do not get the necessary positive reinforcement. While a reward 1.0 may be roughly average on a global scale, they are exceptional by subgroup B standards. However, these responses are not getting as much positive advantage because other, stronger responses exist in \textit{a different subgroup}.

\textbf{Global and Local advantage (\method).} By summing both global and local advantages, we arrive at the following advantage assignment. Within subgroup A, the reward 1.5 responses will receive advantage 1.621, while the reward 1.0 responses receive advantage -1.447. For subgroup B, the reward 1.0 responses will receive advantage 2.017, while the reward 0.0 responses receive advantage -1.811. The sum of the two advantages fixes the issues with Global only: Now, reward 1.0 responses within subgroup B are properly assigned advantage that results in improvement, while the global advantage ties together responses across subgroups, promoting state equivalence.

\subsection{Qualitative \judge{} output examples}
\textbf{Example 1:} From ReasoningJudgeBench, source dataset AIME 2024.
\begin{tcolorbox}[breakable,enhanced, left=-1cm, right=-1cm, top=2pt, bottom=2pt, enlarge top by=0.1cm, enlarge bottom by=0.1cm]
\begin{quote}
\begin{lstlisting}
Question ========================

Solve the following math problem step by step. The last line of your response should be of the form Answer: $Answer (without quotes) where $Answer is the answer to the problem.

Find the sum of all integer bases $b>9$ for which $17_{b}$ is a divisor of $97_{b}$.

Remember to put your answer on its own line after "Answer:".
\end{lstlisting}
\end{quote}
\end{tcolorbox}

\begin{tcolorbox}[breakable,enhanced, left=-1cm, right=-1cm, top=2pt, bottom=2pt, enlarge top by=0.1cm, enlarge bottom by=0.1cm]
\begin{quote}
\begin{lstlisting}
Response A (Incorrect) ========================

To solve this problem, we need to understand what the numbers \(17_b\) and \(97_b\) mean in terms of the base \(b\) and how \(17_b\) might be a divisor of \(97_b\).

### Converting Numbers from Base \(b\)
1. **Convert \(17_b\)**: 
   - In base \(b\), \(17_b\) means \(1 \cdot b + 7\).
   - So, \(17_b = b + 7\).
2. **Convert \(97_b\)**:
   - In base \(b\), \(97_b\) means \(9 \cdot b + 7\).
   - So, \(97_b = 9b + 7\).
   
### Divisibility Condition
We require \(17_b\) to be a divisor of \(97_b\). Thus:
\[97_b = k \times 17_b\]
for some integer \(k\). Substituting the values we found:
\[9b + 7 = k(b + 7)\]

Expanding the right side:
\[9b + 7 = kb + 7k\]

### Solving the Equation
Rearrange the equation:
\[
9b + 7 = kb + 7k \\
9b + 7 = kb + 7k \\
9b - kb = 7k - 7
\]

Factor out \(b\) on the left:
\[
b(9 - k) = 7(k - 1)
\]

### Determine Appropriate Values
Since we need \(b > 9\), try small integer values for \(k\) that could make \(b\) an integer greater than 9. 
\[b = \frac{7(k - 1)}{9 - k}\]

#### Test Values of \(k\):
1. **If \(k = 2\):** \[b = \frac{7(2 - 1)}{9 - 2} = \frac{7}{7} = 1 \quad (\text{Not valid since } b > 9)\]

2. **If \(k = 3\):** \[b = \frac{7(3 - 1)}{9 - 3} = \frac{14}{6} = \frac{7}{3} = 2.\overline{3} \quad (\text{Not valid})\]

3. **If \(k = 4\):** \[b = \frac{7(4 - 1)}{9 - 4} = \frac{21}{5} = 4.2 \quad (\text{Not valid})\]

4. **If \(k = 5\):** \[b = \frac{7(5 - 1)}{9 - 5} = \frac{28}{4} = 7 \quad (\text{Not valid})\]

5. **If \(k = 6\):** \[b = \frac{7(6 - 1)}{9 - 6} = \frac{35}{3} = 11.\overline{6} \quad (\text{Not valid as an integer})\]

6. **If \(k = 7\):** \[b = \frac{7(7 - 1)}{9 - 7} = \frac{42}{2} = 21\]

A valid integer \(b\) is 21. Thus, \(b = 21\) is a base greater than 9 for which \(17_b\) divides \(97_b\) exactly.

### Conclusion
Having checked for divisibility circumstances and calculations, we conclude that the only base \(b > 9\) that satisfies the condition is:

\[b = 21\]

Therefore, the sum of all such bases is:
\[\text{Answer: } 21\]
\end{lstlisting}
\end{quote}
\end{tcolorbox}

\begin{tcolorbox}[breakable,enhanced, left=-1cm, right=-1cm, top=2pt, bottom=2pt, enlarge top by=0.1cm, enlarge bottom by=0.1cm]
\begin{quote}
\begin{lstlisting}
Response B (Correct) ========================

To solve this problem, we first need to convert the numbers \( 17_b \) and \( 97_b \) from base \( b \) to their decimal equivalents:

1. **Convert \( 17_b \) into decimal:** The number \( 17_b \) represents \( 1 \times b + 7 \). Therefore, in decimal, \( 17_b = b + 7 \).

2. **Convert \( 97_b \) into decimal:** The number \( 97_b \) represents \( 9 \times b + 7 \). Therefore, in decimal, \( 97_b = 9b + 7 \).

The problem states that \( 17_b \) (which is \( b + 7 \)) is a divisor of \( 97_b \) (which is \( 9b + 7 \)). 

This implies that: 
\[ 9b + 7 = k(b + 7) \] 
for some integer \( k \). Expanding the right-hand side gives:
\[ 9b + 7 = kb + 7k \]

Subtract \( kb \) from both sides:
\[ 9b + 7 - kb = 7k \]
This simplifies to:
\[ (9 - k)b + 7 = 7k \]
Rearrange to solve for \( b \):
\[ (9 - k)b = 7k - 7 \]
\[ b = \frac{7(k - 1)}{9-k} \]
Since \( b \) must be an integer greater than 9, we find values of \( k \) such that \( 9-k \) divides \( 7(k-1) \) and the resulting \( b \) is an integer greater than 9.

**Test small values of \( k \):**
1. **\( k = 1 \):** \[ b = \frac{7(1-1)}{9-1} = \frac{0}{8} = 0 \,\text{(not valid since } b > 9) \]

2. **\( k = 2 \):** \[ b = \frac{7(2-1)}{9-2} = \frac{7}{7} = 1 \,\text{(not valid since } b > 9) \]

3. **\( k = 3 \):** \[ b = \frac{7(3-1)}{9-3} = \frac{14}{6} \,\text{(not an integer)} \]

4. **\( k = 4 \):** \[ b = \frac{7(4-1)}{9-4} = \frac{21}{5} \,\text{(not an integer)} \]

5. **\( k = 5 \):** \[ b = \frac{7(5-1)}{9-5} = \frac{28}{4} = 7 \,\text{(not valid since } b > 9) \]

6. **\( k = 6 \):** \[ b = \frac{7(6-1)}{9-6} = \frac{35}{3} \,\text{(not an integer)} \]

7. **\( k = 7 \):** \[ b = \frac{7(7-1)}{9-7} = \frac{42}{2} = 21 \,\text{(valid)} \]

8. **\( k = 8 \):** \[ b = \frac{7(8-1)}{9-8} = \frac{49}{1} = 49 \,\text{(valid)} \]

9. **\( k = 9 \):** \[ b = \frac{7(9-1)}{9-9} \] is undefined because this involves division by zero.

The valid integer bases \( b \) for which \( b > 9 \) are \( b = 21 \) and \( b = 49 \). Therefore, their sum is: \[ 21 + 49 = 70 \]

Answer: 70
\end{lstlisting}
\end{quote}
\end{tcolorbox}

\begin{tcolorbox}[breakable,enhanced, left=-1cm, right=-1cm, top=2pt, bottom=2pt, enlarge top by=0.1cm, enlarge bottom by=0.1cm]
\begin{quote}
\begin{lstlisting}
Judge explanation ========================

Explanation: Both Assistant A and Assistant B provided detailed solutions to the problem. However, there is a critical mistake in the final step of Assistant A's response. After correctly identifying that \( b = 21 \) is a valid base, Assistant A incorrectly concludes that this is the only valid base. In contrast, Assistant B correctly identifies both valid bases (\( b = 21 \) and \( b = 49 \)) and calculates their sum correctly.

Assistant B's response is more accurate because it correctly identifies all valid bases that satisfy the given conditions and calculates their sum correctly. Assistant A's response, while detailed, misses the second valid base (\( b = 49 \)) and therefore provides an incorrect final answer.

Verdict: [[B>A]]
\end{lstlisting}
\end{quote}
\end{tcolorbox}

\textbf{Example 2:} From ReasoningJudgeBench, source dataset BIGBench Extra Hard, Disambiguation QA split
\begin{tcolorbox}[breakable,enhanced, left=-1cm, right=-1cm, top=2pt, bottom=2pt, enlarge top by=0.1cm, enlarge bottom by=0.1cm]
\begin{quote}
\begin{lstlisting}
Question ========================

Answer the following multiple choice question. Think step-by-step before outputting your answer. When you output your answer, repeat your answer choice 5 times.
For example, if you select A, output your answer as AAAAA.

Here is a sentence with pronoun(s) whose antecedent(s) can either be derived from the context or is ambiguous.
Sentence: Two friends are travelling in Europe during the summer. One of them proposed to go to London to visit his uncle who lives there. He has not seen members of his family for a while.
Which of the following options correctly explains the antecedent(s) of the pronoun(s)?
(A) The uncle has not seen members of his family for a while.
(B) One of the two friends has not seen members of his family for a while.
(C) It is ambiguous who has not seen members of his family.
(D) One of the two friends is a female.
(E) None of the above.

Output your answer strictly following this format:
Reasoning: Your reasoning here
Answer: Your choice, repeated 5 times
\end{lstlisting}
\end{quote}
\end{tcolorbox}

\begin{tcolorbox}[breakable,enhanced, left=-1cm, right=-1cm, top=2pt, bottom=2pt, enlarge top by=0.1cm, enlarge bottom by=0.1cm]
\begin{quote}
\begin{lstlisting}
Response A (Correct) ========================

Reasoning: Let's analyze the sentence step by step. The sentence describes a situation involving two friends traveling in Europe. One of them proposed going to London to visit "his" uncle, which suggests that the uncle belongs to one of the two friends. The next part of the sentence states, "He has not seen members of his family for a while." The antecedent of "He" could logically be the friend who proposed the visit, implying that this friend has not seen members of his family for a while. However, it could also be interpreted that the uncle has not seen his family for a while, making the antecedent ambiguous. Therefore, without additional context, it is unclear who "he" refers to in this scenario, as both interpretations are plausible. Hence, (C) is the correct choice, indicating that it is ambiguous who has not seen members of his family.

Answer: CCCCC
\end{lstlisting}
\end{quote}
\end{tcolorbox}

\begin{tcolorbox}[breakable,enhanced, left=-1cm, right=-1cm, top=2pt, bottom=2pt, enlarge top by=0.1cm, enlarge bottom by=0.1cm]
\begin{quote}
\begin{lstlisting}
Response B (Correct) ========================

Reasoning: In the given sentence, the pronoun "he" is used in "He has not seen members of his family for a while." To determine the antecedent of "he," we consider the context provided. The sentence starts with "Two friends are travelling...", and later says "One of them proposed to go to London to visit his uncle who lives there." Here, "his uncle" suggests the uncle of one of the friends. We are trying to figure out who "he" refers to in the last part of the sentence. Since two male pronouns ("his" and "he") are used, it's reasonable to think both times it refers to the same person, meaning it refers to the same friend who suggested visiting his uncle. Thus, "he" most likely refers to one of the two friends. Therefore, the pronoun "he" in the final sentence does not introduce ambiguity within the context, and it refers to one of the two friends. 

Answer: BBBBB
\end{lstlisting}
\end{quote}
\end{tcolorbox}

\begin{tcolorbox}[breakable,enhanced, left=-1cm, right=-1cm, top=2pt, bottom=2pt, enlarge top by=0.1cm, enlarge bottom by=0.1cm]
\begin{quote}
\begin{lstlisting}
Judge explanation ========================

Explanation: Both Assistant A and Assistant B provided thoughtful analyses of the sentence to determine the antecedent of the pronoun "he." However, their conclusions differ. Assistant A correctly identifies that the antecedent of "he" is ambiguous because it could refer to either the friend who proposed the visit or the uncle. Assistant B, on the other hand, concludes that "he" refers to one of the two friends, which is a valid interpretation but not the only one, as the sentence does not provide enough context to definitively determine who "he" is. Therefore, Assistant A's answer is more accurate because it acknowledges the ambiguity in the sentence, which is the correct response given the options provided.

Verdict: [[A>B]]
\end{lstlisting}
\end{quote}
\end{tcolorbox}

\section{Prompt templates}\label{app:prompt}
\subsection{J4R prompt templates.}\label{app:prompt:tj}
We present the prompt template used to evaluate \judge{}, \judgeins{}, and CompassJudger below. Originally, CompassJudger evaluates on JudgeBench using the ArenaHard prompt, which first tasks the judge with \textit{solving the original problem} prior to evaluation, which can be viewed as a form of inference-time compute scaling. Our prompt only prompts the judge with evaluation, which we find boosts the 7B model performance but leads to slight degradations for the 14B and 32B versions, as noted in the main text. Furthermore, we are unable to reproduce officially reported CompassJudger numbers, the evaluation of which was run by JudgeBench. For example, JudgeBench reports that CompassJudger-7B achieves 46.00\%. Using JudgeBench's own scripts, we obtain a score of 34.43\%, whereas with our own implementation, we obtain a score of 45.71\%. As a result, we report deviations with respect to our reproduced numbers: 45.71\% for 7B, 54.29\% for 14B, and 59.14\% for 32B. 

\begin{tcolorbox}[breakable,enhanced, left=-1cm, right=-1cm, top=2pt, bottom=2pt, enlarge top by=0.1cm, enlarge bottom by=0.1cm]
\begin{quote}
\begin{lstlisting}
### System Prompt
Please act as an impartial judge and evaluate the quality of the responses provided by two AI assistants to the user prompt displayed below. You will be given assistant A's answer and assistant B's answer. Your job is to evaluate which assistant's answer is better.

When evaluating the assistants' answers, identify any mistakes or inaccurate information. Focus on the content each response, and select the response that arrives at the correct response and with the most sound and logical reasoning. Avoid any biases, such as order of responses, length, or stylistic elements like formatting.

Before providing an your final verdict, think through the judging process and output your thoughts as an explanation

After providing your explanation, you must output only one of the following choices as your final verdict with a label:

1. Assistant A is better: [[A>B]]
2. Assistant B is better: [[B>A]]

Use the following template:
Explanation: Your detailed thought process as an explanation.
Verdict: [[A>B]] or [[B>A]].


### User Prompt
<|User Prompt|>
{question}

<|The Start of Assistant A's Answer|>
{response_a}
<|The End of Assistant A's Answer|>

<|The Start of Assistant B's Answer|>
{response_b}
<|The End of Assistant B's Answer|>
\end{lstlisting}
\end{quote}
\end{tcolorbox}

\subsection{Benchmark generation prompts}
Below, we provide two sample prompts for generating model responses to (1) multiple choice questions and (2) math questions when constructing \benchmark{}.

\begin{tcolorbox}[breakable,enhanced, left=-1cm, right=-1cm, top=2pt, bottom=2pt, enlarge top by=0.1cm, enlarge bottom by=0.1cm]
\begin{quote}
\begin{lstlisting}
### Multiple choice prompt

Answer the following multiple choice question. Think step-by-step before outputting your answer. When you output your answer, repeat your answer choice 5 times.
For example, if you select A, output your answer as AAAAA.

{optional context}
Question: {question}

{choices}

Output your answer strictly following this format:
Reasoning: Your reasoning here
Answer: Your choice, repeated 5 times
\end{lstlisting}
\end{quote}
\end{tcolorbox}

\begin{tcolorbox}[breakable,enhanced, left=-1cm, right=-1cm, top=2pt, bottom=2pt, enlarge top by=0.1cm, enlarge bottom by=0.1cm]
\begin{quote}
\begin{lstlisting}
### Math prompt

Solve the following math problem step by step. The last line of your response should be of the form Answer: $Answer (without quotes) where $Answer is the answer to the problem.

{question}

Remember to put your answer on its own line after "Answer:".
\end{lstlisting}
\end{quote}
\end{tcolorbox}

\end{document}